\documentclass{article}

\usepackage[main, final]{neurips_2026}

\usepackage[utf8]{inputenc} 
\usepackage[T1]{fontenc}    
\usepackage{hyperref}       
\usepackage{url}            
\usepackage{booktabs}       
\usepackage{amsfonts}       
\usepackage{nicefrac}       
\usepackage{microtype}      
\usepackage{xcolor}         
\usepackage[small]{caption}
\usepackage{graphicx}
\usepackage{amsmath}
\usepackage{wrapfig}
\usepackage{tabularx}
\usepackage{amsthm}
\usepackage{amssymb}
\usepackage{booktabs}
\usepackage{algorithm}
\usepackage{algorithmic}
\usepackage{booktabs}
\usepackage{pifont}
\usepackage{colortbl}
\usepackage[most]{tcolorbox} 
\usepackage{enumitem}
\usepackage{longtable}
\newcommand{\sysname}{TTCS}
\definecolor{rowblue}{HTML}{E6F3FF} 
\definecolor{textred}{HTML}{FF3333} 
\definecolor{textgreen}{HTML}{009900} 
\title{\textit{TTCS}: Test-Time Curriculum Synthesis for Self-Evolving}

\author{%
  Chengyi Yang\textsuperscript{1,2}, 
  Zhishang Xiang\textsuperscript{1,2}, 
  Yunbo Tang\textsuperscript{1,2}, 
  Zongpei Teng\textsuperscript{1,2}, \\
  \textbf{Chengsong Huang\textsuperscript{3}, 
  Fei Long\textsuperscript{1}\thanks{Corresponding authors}\;,
  Yuhan Liu\textsuperscript{4}\footnotemark[1]\;,
  Jinsong Su\textsuperscript{1,2}\footnotemark[1]} \\
  \textsuperscript{1}Xiamen University \quad  \textsuperscript{2}Key Lab of Digital Protection and Intelligent Processing of\\ Intangible Cultural Heritage of Fujian-Taiwan (XMU), Ministry of Culture and Tourism \\
  \textsuperscript{3}Washington University in St. Louis \quad
  \textsuperscript{4}Renmin University of China \\
  \texttt{yangchengyi@stu.xmu.edu.cn} \quad \texttt{yuhan.liu@ruc.edu.cn} \quad\texttt{jssu@xmu.edu.cn}
}

\begin{document}

\maketitle


\begin{abstract}
Test-Time Training offers a promising way to improve the reasoning ability of large language models (LLMs) by adapting the model using only the test questions. However, existing methods struggle with difficult reasoning problems for two reasons: raw test questions are often too difficult to yield high-quality pseudo-labels, and the limited size of test sets makes continuous online updates prone to instability. To address these limitations, we propose \textbf{\sysname{}}, a co-evolving test-time training framework. Specifically, \sysname{} initializes two policies from the same pretrained model: a question \textit{synthesizer} and a reasoning \textit{solver}. These policies evolve through iterative optimization: the synthesizer generates progressively challenging question variants conditioned on the test questions, creating a structured curriculum tailored to the solver's current capability, while the solver updates itself using self-consistency rewards computed from multiple sampled responses on both original test and synthetic questions. Crucially, the solver's feedback guides the synthesizer to generate questions aligned with the model's current capability, and the generated question variants in turn stabilize the solver's test-time training. Experiments show that \sysname{} consistently strengthens the reasoning ability on challenging mathematical benchmarks and transfers to general-domain tasks across different LLM backbones, highlighting a scalable path towards dynamically constructing test-time curricula for self-evolving. Our code and implementation details are available at \textcolor{blue}{\url{https://github.com/XMUDeepLIT/TTCS}}.
\end{abstract}

\section{Introduction}

\begin{figure}[!htbp] 
    \centering
    
    \includegraphics[width=.80\linewidth]{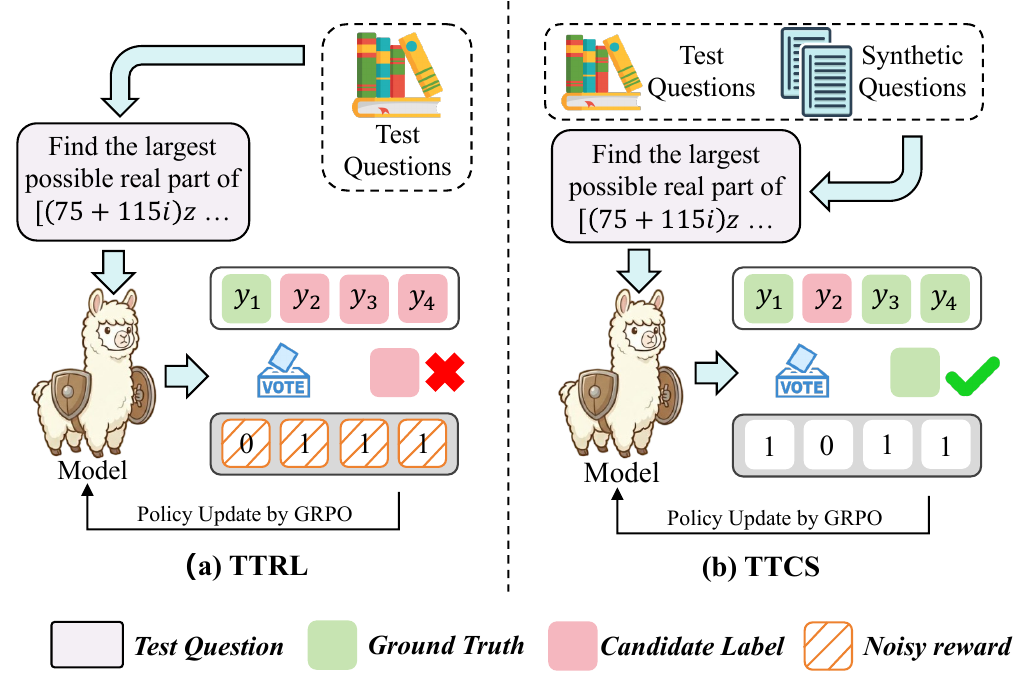} 
    \caption{Comparison of TTRL and our TTCS: (a) When applied to difficult test questions such as AIME24, TTRL suffers from noisy rewards caused by incorrect majority voting consensus. (b) TTCS synthesizes tractable variants to ensure valid pseudo-labels, providing reliable supervision for stable self-evolution.}
    \label{fig:overview}
\end{figure}

Large language models (LLMs) have evolved from passive text generators into increasingly autonomous agents capable of planning, acting, and reasoning over complex tasks~\cite{grattafiori2024llama3herdmodels,kimiteam2025kimik2openagentic,ma2025generalreasoneradvancingllmreasoning,jin2025searchr1trainingllmsreason,zhang2025weaving}. This progress has been largely driven by reinforcement learning with verifiable rewards (RLVR)~\cite{shao2024deepseekmathpushinglimitsmathematical,yu2025dapoopensourcellmreinforcement,min2026orchestratingtokenssequencesdynamic,liu2026healingentropycollapseenhancing,liu2026specrlacceleratingonpolicyreinforcement}, which has yielded significant improvements on challenging mathematical and scientific benchmarks~\cite{deepseekai2025deepseekr1incentivizingreasoningcapability,yang2025qwen3technicalreport}. By leveraging verifiable outcomes, RLVR allows models to iteratively refine their strategies through outcome-based feedback. However, this paradigm faces a critical scalability bottleneck: it heavily depends on extensive, high-quality supervised data~\cite{silver2025welcome}. 

To address this limitation, recent studies~\cite{chen2024selfplayfinetuningconvertsweak,zhao2025absolutezeroreinforcedselfplay,liu2025spiceselfplaycorpusenvironments,fang2025serlselfplayreinforcementlearning,huang2025rzeroselfevolvingreasoningllm,zhang2025viper} have shifted toward \textit{Self-Evolving}. These approaches empower models to improve autonomously through self-generated supervision and environmental interactions, thereby reducing reliance on external human labels. A representative work of this paradigm is Test-Time 
Training~\cite{sun2020testtimetrainingselfsupervisiongeneralization}, particularly its reinforcement learning variant known as test-time reinforcement learning (TTRL)~\cite{zuo2025ttrltesttimereinforcementlearning}. Originally designed to mitigate distribution shifts, TTRL
enables dynamic parameter adaptation on unlabeled test instances by 
optimizing a self-supervised objective, which is typically constructed via majority voting~\cite{wang2023selfconsistencyimproveschainthought}. However,  as illustrated in Figure~\ref{fig:overview}(a), when applied 
to challenging reasoning tasks, TTRL encounters two primary challenges: 
(i) \textbf{Unreliable Pseudo-labels.} 
For difficult questions such as AIME24~\cite{li2025limrrlscaling}, the majority of sampled responses are often incorrect. Majority voting therefore converges to a wrong consensus, producing systematically noisy reward signals that actively misguide the policy update~\cite{zhao2025majorityrightrltraining}. 
(ii) \textbf{Optimization lacks learnable samples.} 
TTRL operates directly on a small set of extremely challenging test questions. As shown in Figure~\ref{fig:overview}(a), these questions lie far beyond the model's current capability. Without intermediate variants to bridge the gap, the learning process becomes steep and often unclimbable~\cite{briesch2024largelanguagemodelssuffer}.

To overcome these challenges, we broaden the role of test-time scaling from repeatedly solving a fixed question to constructing problem-specific experience for self-evolving. 
Conventional test-time scaling spends additional computation on the same test question, using repeated sampling, search, or majority voting to obtain a better final answer~\cite{wang2023selfconsistencyimproveschainthought,snell2024scalingllmtesttimecompute,liu20251bllmsurpass405b}. 
This improves the chance of finding a correct answer, but it does not make the original question easier to learn. 
When the question is beyond the model's current capability, repeated attempts often produce incorrect or inconsistent answers, so the resulting self-supervised reward provides little reliable guidance for updating the model.
We therefore construct a local problem series around the target question, enabling the model to learn from related, more tractable variants before returning to the original hard problem. 
In this sense, the test question is not merely an input to be answered, but an anchor for building a local training distribution at inference time.


In this paper, we propose \textbf{\sysname{}}  (\textbf{T}est-\textbf{T}ime \textbf{C}urriculum \textbf{S}ynthesis for Self-Evolving), a co-evolving test-time training framework that couples \emph{capability-aware synthesizer} with \emph{online self-evolving solver}. 
\sysname{} instantiates two agents initialized from the same pretrained model: a \textbf{Synthesizer} that, given a test question, generates curriculum questions variants that preserve the underlying reasoning structure while varying surface realizations; and a \textbf{Solver} that performs online training on a mixture of test questions and synthetic questions. 
Crucially, the two agents co-evolve in an iterative loop: the solver serves as an implicit judge of each synthesized question quality, providing a capability-aligned signal that trains the synthesizer to propose auxiliary questions near the solver's capability frontier, while the solver is updated using self-supervised rewards derived from its own sampled responses~\cite{zuo2025ttrltesttimereinforcementlearning,wang2023selfconsistencyimproveschainthought}. 
Both agents are optimized online via Group Relative Policy Optimization (GRPO)~\cite{shao2024deepseekmathpushinglimitsmathematical}, enabling stable self-evolving under label-free test-time constraints. Generally, our main contributions are summarized as follows:

\begin{itemize}
\setlength{\itemsep}{0pt}
\setlength{\parsep}{0pt}
\setlength{\parskip}{0pt}
    \item We identify limitations in existing test-time training for complex reasoning, including unreliable pseudo-labels and the lack of learnable samples, which collectively hinder effective optimization on difficult test questions in practice.
    \item We propose \textbf{\sysname{}}, a co-evolving test-time training framework. TTCS couples a capability-aware synthesizer with an online solver. Through iterative GRPO, the synthesizer dynamically constructs a problem-centered curriculum aligned with the solver's capability frontier, thereby transforming noisy feedback into a reliable pathway for self-evolution.
    \item Extensive experiments show that our approach achieves strong performance and generalizes well across both mathematical and general reasoning benchmarks.
\end{itemize}

\section{Related Work}
The related works to ours mainly include the following two lines of studies:

\textbf{Self Evolving for LLMs.} Self Evolving has been considered as a paradigm for augmenting LLMs' reasoning ability without extra human supervision. In this regard, pioneering studies~\cite{huang2022largelanguagemodelsselfimprove,selfinstruct} demonstrate that LLMs can achieve meaningful self-improvement by fine-tuning on their own high-confidence reasoning trajectories. Following this, SPIN~\cite{chen2024selfplayfinetuningconvertsweak} introduces iterative self-play fine-tuning. To address the degenerative feedback in single-model self-play, later role-specialized frameworks~\cite{lin2025learningsolveverifyselfplay,chen2025spcevolvingselfplaycritic} adopt co-evolution as a core principle, which scales effectively in low-data settings~\cite{fang2025serlselfplayreinforcementlearning,shafayat2025largereasoningmodelsselftrain}. Recent studies shift toward dynamic self-challenging~\cite{zhou2025selfchallenginglanguagemodelagents,liang2025swsselfawareweaknessdrivenproblem} and unsupervised post-training~\cite{wei2025sftsecondrlupt}. This progress marks a transition from imitating external supervision~\cite{yu2025guided} to autonomous self-correction based on intrinsic verifiability, leading to zero-data systems~\cite{huang2025rzeroselfevolvingreasoningllm,zhao2025absolutezeroreinforcedselfplay,he2025visplay}. However, as noted by~\cite{shumailov2024ai}, recursive training on self-generated data carries the risk of model collapse, underscoring the need for high-quality data synthesis mechanisms.

\textbf{Test-Time Training (TTT).} This paradigm dynamically adapts model parameters during inference via self-supervision~\cite{akyürek2025surprisingeffectivenesstesttimetraining}. A key development is Test-Time Reinforcement Learning (TTRL)~\cite{zuo2025ttrltesttimereinforcementlearning}, which applies RLVR to unlabeled test data by deriving pseudo-labels from multi-sampling, allowing LLMs to self-improve during inference. While AlphaProof~\cite{hubert2025olympiad} has demonstrated the substantial potential of TTT in tackling Olympiad-level mathematical reasoning, it predominantly relies on stronger LLM~\cite{geminiteam2024gemini15unlockingmultimodal,comanici2025gemini25pushingfrontier} for data synthesis, limiting the scope of fully autonomous self-evolving. To overcome this limitation, we introduce \sysname{}, a fully autonomous framework that eliminates the need for external supervision and stronger teacher models.



\section{Preliminary}
\label{subsec:preliminary}
This section reviews two key methodologies: Group Relative Policy Optimization~\cite{shao2024deepseekmathpushinglimitsmathematical} as a core optimization algorithm, and the Test-Time Training paradigm~\cite{zuo2025ttrltesttimereinforcementlearning} for label-free inference adaptation.

\textbf{Group Relative Policy Optimization (GRPO).}\label{preliminary:GRPO} Let $\mathcal{X}$ and $\mathcal{Y}$ denote the input question space and output response space, respectively. We denote the LLM as a policy $\pi_\theta$ parameterized by $\theta$, which generates a response $y \in \mathcal{Y}$ given an input $x \in \mathcal{X}$ according to the conditional probability $\pi_\theta(y\mid x)$.
In the standard RLVR setting, we utilize a dataset $\mathcal{D}=\{(x, y^\ast)\}$, where $y^\ast$ represents the ground truth. This allows us to define an outcome-based reward function $\mathcal{R}(y, y^\ast): \mathcal{Y} \times \mathcal{Y} \to \{0, 1\}$, which evaluates the correctness of the generated response.
The optimization goal is to maximize the expected reward:
\begin{equation}
    \theta^\ast = \arg\max_{\theta}\ \mathbb{E}_{y \sim \pi_\theta(\cdot\mid x)} \big[\mathcal{R}(y,y^\ast)\big].
\end{equation}
To optimize this objective efficiently, GRPO has emerged as a widely adopted optimization algorithm. Specifically, for each question $x$, the model samples a group of outputs $\{o_i\}_{i=1}^{G}$ from the current policy $\pi_{\theta_{old}}$. The advantage $A_i$ for the $i$-th output $o_i$ is computed by normalizing the rewards with respect to the group statistics:
\begin{equation}
    A_i = \frac{r_i - \text{mean}(\mathbf{r})}{\text{std}(\mathbf{r}) + \epsilon},
\end{equation}
where $\mathbf{r} = \{r_1, \dots, r_G\}$ represents the group rewards and $\epsilon$ is a small constant for numerical stability. 
The policy is updated by maximizing the surrogate objective $\mathcal{J}_{GRPO}(\theta)$ as follows:
\begin{equation}
\begin{aligned}
    \mathcal{J}_{GRPO}(\theta) = 
    \mathbb{E}_{\substack{x \sim \mathcal{D}, \{o_i\}_{i=1}^G \sim \pi_{\theta_{old}}(\cdot|x)}} \Bigg[ \frac{1}{G} \sum_{i=1}^G \Bigg( \min \bigg( \frac{\pi_\theta(o_i|x)}{\pi_{\theta_{old}}(o_i|x)} A_i, \\
     \text{clip}\left( \frac{\pi_\theta(o_i|x)}{\pi_{\theta_{old}}(o_i|x)}, 1-\epsilon, 1+\epsilon \right) A_i \bigg) 
    -\beta \mathbb{D}_\text{KL}(\pi_\theta || \pi_{old}) \Bigg) \Bigg],
\end{aligned}
\end{equation}
where $\epsilon$ and $\beta$ are hyper-parameters. The $\text{clip}(\cdot)$ function ensures stable updates within a trust region, while the KL term $\mathbb{D}_{\text{KL}}$ acts as a regularizer to prevent excessive deviation from the previous policy model.

\textbf{Test-Time Training (TTT).}\label{preliminary:TTT} Let the test dataset $\mathcal{D}_{test}=\{(x_\text{test}, y_\text{test})\}$, where the $x_\text{test}$ and $y_\text{test}$ are the test question and the ground-truth, respectively. TTT is a paradigm designed to mitigate the distribution shift between training and testing environments~\cite{zuo2025ttrltesttimereinforcementlearning}. Unlike standard inference with fixed parameters, TTT enables the model to dynamically adapt its parameters only using test questions.
Without access to ground-truth, TTT relies on self-generated supervision. Formally, starting from a pretrained model, TTT seeks to find adapted parameters $\theta_{\mathrm{ttt}}^*$ by maximizing a label-free objective over the test questions:
\begin{equation}
    \theta_{\mathrm{TTT}}^* = \arg\max_{\theta}\ \mathbb{E}_{y \sim \pi_\theta(\cdot\mid x_{\text{test}})} \big[\mathcal{R}_{\mathrm{TTT}}(y, \hat{y}^*)\big],
\end{equation}
where $\hat{y}^*$ is the pseudo-label derived via the majority voting~\cite{wang2023selfconsistencyimproveschainthought}, which selects the most frequent response among the candidates, and $\mathcal{R}_{\mathrm{TTT}}(y, \hat{y}^*)$ denotes the outcome-based reward measuring the alignment between the response $y$ and the consensus $\hat{y}^*$.

\section{Test-Time Curriculum Synthesis}
\label{sec:method}
\begin{figure*}[t] 
    \centering
    
    \includegraphics[width=0.95\textwidth]{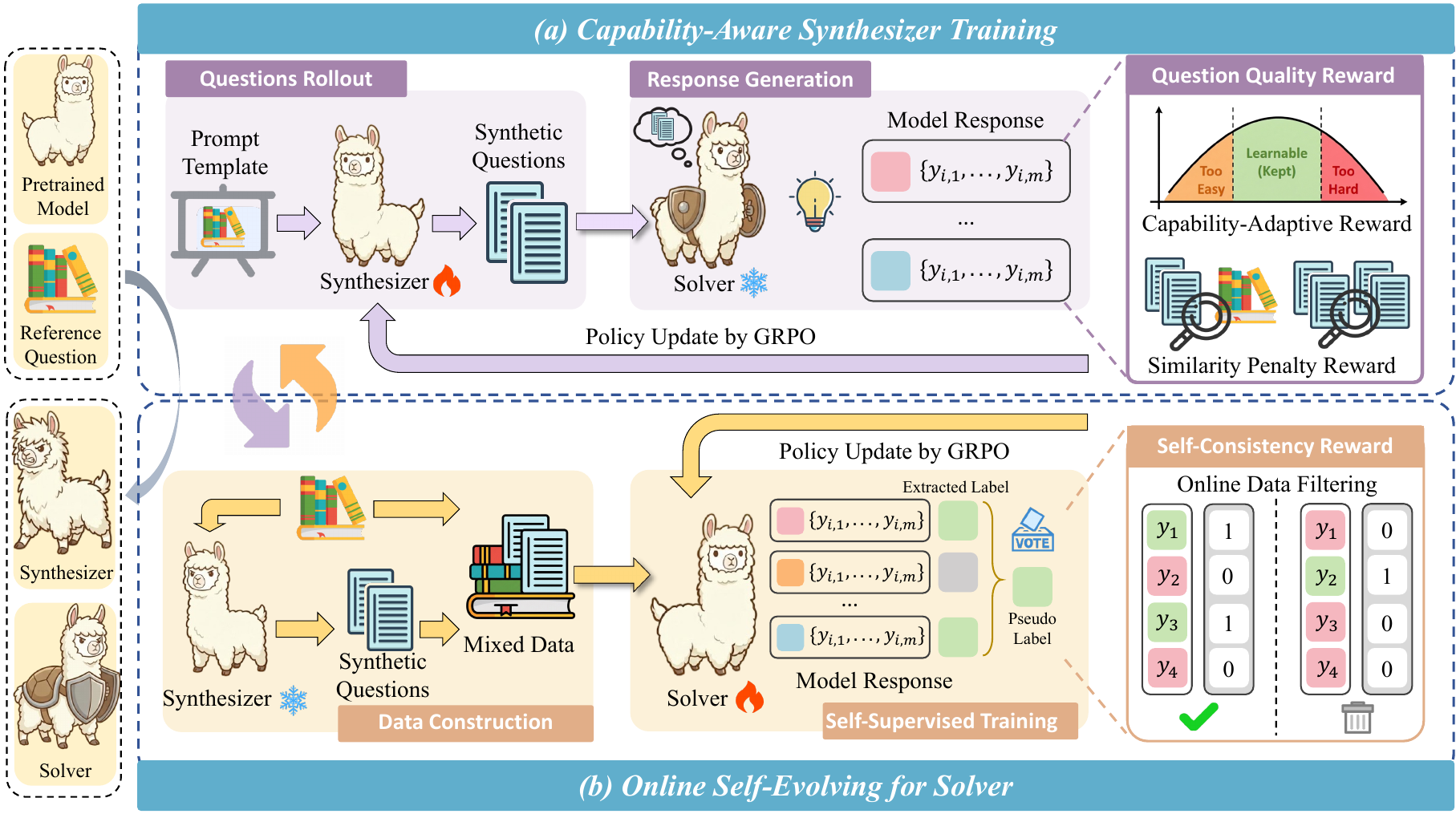} 
    \caption{Overview of \sysname{}, a co-evolving test-time training framework.
(\textbf{a}) \emph{Synthesizer training}: the synthesizer first rollout synthetic questions conditioned on test questions with the prompt template, and then is optimized via GRPO with the question quality reward.
(\textbf{b}) \emph{Solver training}: the solver performs online self-evolving on a mixture of test and synthetic questions with self-supervised rewards, updated via GRPO using the self-consistency reward.}
    \label{fig:framework}
\end{figure*}

In this section, we present \textbf{T}est-\textbf{T}ime \textbf{C}urriculum \textbf{S}ynthesis (\textbf{\sysname{}}), a co-evolving test-time training framework built on an iterative GRPO (Section~\ref{subsec:preliminary}) optimization loop. As shown in Figure~\ref{fig:framework}, \sysname{} consists of two agents: a \textbf{Synthesizer} policy $\pi_{\phi}$ and a \textbf{Solver} policy $\pi_{\theta}$, both initialized from the same pretrained model. At each iteration, the synthesizer generates curriculum variants for each test question, and is rewarded to preserve the reasoning structure of each test question while staying near the solver's current capability frontier(Section~\ref{subsec:synthesizer}). The solver, in turn, performs online self-evolving on a mixture of synthetic questions and test questions, guided by self-consistency rewards(Section~\ref{subsec:solver}). Crucially, the two agents co-evolve in a closed loop: the solver's current performance provides a capability-aware training signal that shapes the synthesizer's generation distribution, while the synthesizer continuously supplies fresh, question-centered variants that stabilize the solver's test-time training. The algorithm description of our framework is shown in Appendix~\ref{appendix:TTCS}

\subsection{Capability-Aware Synthesizer Training}
\label{subsec:synthesizer}
Since difficult and limited test questions often provide weak and unreliable pseudo-labels at test-time training~\cite{wei2025sftsecondrlupt}, the structured and localized curriculum of variants around each test question is vital~\cite{hubert2025olympiad}, allowing the solver to learn from simpler, related variants rather than from the overly difficult raw test questions alone. We realize this through two components: (i) \textit{Test Questions Guided Synthesis}, which preserves the important reasoning structure of the test question while varying the surface form to produce diverse variants; and (ii) \textit{Question Quality Reward}, which favors synthetic questions that the solver can partially solve but not trivially solve, making them most informative for driving test-time self-evolving.

\paragraph{Test Questions Guided Synthesis.}
At the $t$-th iteration, the synthesizer generates $M$ auxiliary questions $\{x^\prime_i\}_{i=1}^{M}$ for each RL rollout group conditioned on $x_{\mathrm{test}}$ with a well-designed prompt template (see Appendix~\ref{appendix:prompt}) as follows:
\begin{equation}
    \{x^\prime_i\}_{i=1}^{M} \sim \pi_{\phi}^{t}(\cdot \mid x_{\mathrm{test}}).
\end{equation}
Each synthetic question $x^\prime_i$ preserves the underlying reasoning structure of $x_{\mathrm{test}}$ to maintain task relevance, while differing in surface realization through changes in problem objects, settings, or constraint types. This allows the synthetic questions to provide focused training signals that are aligned with the reference questions in distribution.

\paragraph{Question Quality Reward.} To facilitate the co-evolution of the synthesizer and the solver, we employ the solver as an online assessor to assign a composite reward to each synthetic question in rollout stage. This reward is designed to reflect two complementary objectives: 


\textbf{Capability-Adaptive Reward.} We first quantify the difficulty of each synthetic question $x^\prime_i$ related to the solver's current policy $\pi_\theta^t$. By sampling $K$ reasoning responses $\{y_{i,k}\}_{k=1}^{K} \sim \pi_{\theta}^t( \cdot \mid x^\prime_i)$ and computing the majority vote $\hat{y}_{\mathrm{i}}$ as implemented in~\cite{wang2023selfconsistencyimproveschainthought}, we define the self-consistency score to measure the difficulty of $x^\prime_i$ as:
\begin{equation}
\label{eq:consistency}
    s(x^\prime_i) = \frac{1}{K} \sum_{k=1}^{K} \mathbb{I}\big[y_{i,k} = \hat{y}_{i}\big].
\end{equation}
To maximize the training efficiency, the synthesizer should target the solver's capability frontier, which consists of problems that are neither too easy (consistently solved) nor too hard (consistently failed). Since the variance of the generated responses distribution peaks when the self-consistency score $s(x^\prime_i)\approx 0.5$, we formulate a variance-driven capability-adaptive reward~\footnote{The detailed theoretical analysis of this variance-driven reward is shown in Appendix ~\ref{appendix:theory_reward}} to prioritize this as follows:
\begin{equation}
    \mathcal{R}_{\mathrm{cap}}(x^\prime_i) = \big(4\, s(x^\prime_i) (1 - s(x^\prime_i))\big)^{\gamma},
\end{equation}
where $\gamma$ controls the curvature, assigning peak rewards to questions at the capability boundary of the solver $\pi_{\theta}^t$.

\textbf{Similarity Penalty Reward.}\label{method:similarity_penalty} To ensure novelty and mitigate model collapse, we introduce reward penalty terms that discourage trivial copying from the test questions $x_{\mathrm{test}}$ and redundancy within the other synthetic questions $\{x^\prime_{i^{\prime}}\}_{i^{\prime}=1, i^{\prime}\neq i}^{M}$ in the same rollout group.  Specifically, $\mathcal{R}_{\mathrm{ref}}(x^\prime_i, x_{\mathrm{test}})$ penalizes near-duplicate paraphrases based on rule-based similarity, such as the edit distance between the synthetic question and the test one (see Appendix~\ref{appendix:ref_panalty}), while $\mathcal{R}_{\mathrm{group}}(x^\prime_i, \{x^\prime_{i^{\prime}}\}_{i^{\prime}=1, i^{\prime}\neq i}^{M})$ penalizes redundancy among generated samples to encourage exploration (see Appendix~\ref{appendix:group_panalty}). The overall similarity penalty reward is computed as a weighted sum of these two components:
\begin{equation}
    \mathcal{R}_{\mathrm{sim}}(x^\prime_i) = \lambda_1 \mathcal{R}_{\mathrm{ref}}(x^\prime_i, x_{\mathrm{test}}) + \lambda_2 \mathcal{R}_{\mathrm{group}}(x^\prime_i, \{x^\prime_{i^{\prime}}\}_{i^{\prime}=1, i^{\prime}\neq i}^{M}),
\end{equation}
where $\lambda_1$ and $\lambda_2$ are scalar coefficient that control the relative strength of the reference-based and group-level penalties, respectively.

\textbf{Final Reward Formulation.} By combining the capability objective with diversity constraints, we define the final reward as follows to guide the training process:
\begin{equation}
    \mathcal{R}(x^\prime_i) = \mathbb{I}_{\text{valid}}(x^\prime_i) \cdot \max\left(0, \mathcal{R}_{\mathrm{cap}}(x^\prime_i) - \mathcal{R}_{\mathrm{sim}}(x^\prime_i)\right),
\end{equation}
where $\mathbb{I}_{\text{valid}}(x^{\prime}_i)$ is an indicator that enforces basic format compliance (e.g., the synthesized question is properly wrapped in \texttt{<question>} \,$\cdots$\, \texttt{</question>}). The synthesizer $\pi_{\phi}^{t}$ is then updated with GRPO using $\mathcal{R}(x^{\prime}_i)$, so that its generation distribution progressively shifts toward capability-aligned and diverse variants.

\subsection{Online Self-Evolving for Solver}
\label{subsec:solver}
Parallel to the synthesizer's evolution, the solver is iteratively refined to adapt to the increasingly challenging curriculum. While the synthesizer acts as a curriculum designer to near the capability frontier, the solver functions as an adaptive learner, continuously updating its policy to master these synthetic challenges.


\textbf{Training Data Construction at Test Time.} To balance domain grounding with capability exploration, at iteration $t$, we construct training data by augmenting sampled test questions $x_\mathrm{test}$ with diverse variants $\{x^\prime_i\}_{i=1}^{N}$ generated by the synthesizer $\pi_{\phi}^{t-1}$. The resulting mixed training data is formulated as $\mathcal{B}^{t}_{\text{train}} = \mathcal{B}_{\text{test}}^t \cup \mathcal{B}_{\text{syn}}^{t}$, where $\mathcal{B}_{\text{test}}^t$ represents the test questions set and $\mathcal{B}_{\text{syn}}^{t}$ denotes the corresponding synthetic questions set. Notably, since the test set $\mathcal{D}_{\text{test}}$ is typically small, we repeatedly sample test questions across iterations. This re-sampling strategy controls the ratio between test and synthetic data, preventing the training distribution from being dominated by self-generated samples, which could otherwise lead to model collapse in practice~\cite{shumailov2024ai}.

\textbf{Self-Consistency Reward for Solver.} As described in Section~\ref{preliminary:TTT}, we employ the majority voting mechanism to obtain the pseudo-labels. Specifically, given a training question $x \in \mathcal{B}^{t}_{\text{train}}$, the solver $\pi_\theta^t$ first generates multiple reasoning responses via repeated high temperature sampling:
\begin{equation}
    \{y_i\}_{i=1}^{G} \sim \pi_{\theta}^{t}(\cdot \mid x).
\end{equation}
We then aggregate these responses to derive a consensus prediction $\hat{y}\ast$ , which serves as a pseudo-label. A binary outcome-based reward $R_{TTT}(y_i, \hat{y}^\ast)$ is assigned to each response $y_i$ using its agreement with the consensus.
Based on these rewards, we calculate the sample consistency score $s(x) = \frac{1}{G} \sum_{i=1}^{G} R_{TTT}(y_i, y^{*})$. Following the strategy in DAPO~\cite{yu2025dapoopensourcellmreinforcement}, we implement an \emph{online data filtering} mechanism that retains only samples satisfying $\lvert s(x) - 0.5 \rvert \le \delta$ to ensure the quality of training data used for self-evolving. Together, this online self-evolving procedure enables the solver to continuously refine its policy using only unlabeled test-time data and self-generated supervision.

\section{Experiment}
\subsection{Experimental Setting}
\label{sec:exp-setup}

\begin{table*}[t]
\small
\centering

\renewcommand{\arraystretch}{1.1}
\setlength{\tabcolsep}{2pt}
\begin{tabular}{lccccccc}
\toprule
\textbf{Method} & \textbf{AIME 2024} & \textbf{AIME 2025} & \textbf{AMC23} & \textbf{MATH-500} & \textbf{Minerva} & \textbf{OlympiadBench} & \textbf{AVG} \\
\midrule

\multicolumn{8}{c}{\cellcolor[HTML]{EFEFEF}\textbf{Qwen2.5-Math-1.5B}} \\
\midrule
Pretrained model~\cite{qwen2025qwen25technicalreport} & 7.10 & 4.20 & 27.50 & 33.20 & 9.60 & 22.20 & 17.30 \\
Self-Consistency~\cite{wang2023selfconsistencyimproveschainthought} & 13.30 & 10.00 & 50.00 & 49.80 & 10.70 & 31.90 & 27.62 \\
R-Zero~\cite{huang2025rzeroselfevolvingreasoningllm} & 10.00 & 4.58 & 47.50 & 66.20 & 30.88 & 31.01 & 31.70 \\
TTRL~\cite{zuo2025ttrltesttimereinforcementlearning} & 13.23 & 9.38 & 55.00 & 71.20 & 34.93 & 35.61 & 36.56 \\
\rowcolor{rowblue} \textbf{TTCS (Ours)} & \textbf{19.79} & \textbf{13.33} & \textbf{62.50} & \textbf{76.80} & \textbf{40.44} & \textbf{36.05} & \textbf{41.49} \\
\midrule

\multicolumn{8}{c}{\cellcolor[HTML]{EFEFEF}\textbf{Qwen2.5-Math-7B}} \\
\midrule
Pretrained model~\cite{qwen2025qwen25technicalreport} & 12.90 & 7.90 & 45.00 & 52.80 & 18.80 & 18.70 & 26.02 \\
Self-Consistency~\cite{wang2023selfconsistencyimproveschainthought} & 20.00 & 13.30 & 52.50 & 62.20 & 22.10 & 22.80 & 32.15 \\
R-Zero~\cite{huang2025rzeroselfevolvingreasoningllm} & 18.13 & 7.81 & 65.00 & 78.60 & 43.38 & 39.47 & 42.07 \\
TTRL~\cite{zuo2025ttrltesttimereinforcementlearning} & 35.52 & 14.06 & 67.50 & 83.40 & 49.26 & 40.80 & 48.42 \\
\rowcolor{rowblue} \textbf{TTCS (Ours)} & \textbf{37.19} & \textbf{19.90} & \textbf{75.00} & \textbf{84.60} & \textbf{53.31} & \textbf{45.25} & \textbf{52.54} \\
\midrule

\multicolumn{8}{c}{\cellcolor[HTML]{EFEFEF}\textbf{Qwen3-4B-Base}} \\
\midrule
Pretrained model~\cite{yang2025qwen3technicalreport} & 12.10 & 5.40 & 45.00 & 72.40 & 32.70 & 39.90 & 34.58 \\
Self-Consistency~\cite{wang2023selfconsistencyimproveschainthought} & 20.00 & 10.00 & 57.50 & 79.60 & 41.20 & 44.10 & 42.07 \\
R-Zero~\cite{huang2025rzeroselfevolvingreasoningllm} & 11.35 & 8.65 & 55.00 & 76.20 & 45.96 & 42.73 & 39.98 \\
TTRL~\cite{zuo2025ttrltesttimereinforcementlearning} & 16.67 & 17.81 & 57.50 & 80.40 & 45.96 & 43.18 & 43.59 \\
\rowcolor{rowblue} \textbf{TTCS (Ours)} & \textbf{25.00} & \textbf{19.58} & \textbf{60.00} & \textbf{81.80} & \textbf{52.21} & \textbf{44.66} & \textbf{47.21} \\

\bottomrule
\end{tabular}
\caption{Comparison of our \textbf{\sysname{}} framework against other baselines on mathematical benchmarks. Best results are highlighted in bold.}
\label{tab:main_results}
\end{table*}

\textbf{Datasets and Evaluation.}
We apply \textbf{\sysname{}} to each benchmark individually and then evaluate to demonstrate its effectiveness. \textbf{(1) Competition-Level Mathematical Benchmarks:} We employ the AMC23, AIME24, and AIME25 as a rigorous testbed for advanced reasoning ability~\cite{li2025limrrlscaling}. \textbf{(2) Fundamental Mathematical Benchmarks:} Complementarily, we include MATH-500~\cite{MATH-500}, Minerva~\cite{Minerva}, and OlympiadBench~\cite{OlympiadBench} to assess fundamental mathematical proficiency across diverse problem types (See Appendix~\ref{appendix:datasets}). Following the setting of prior works~\cite{zeng2025simplerlzooinvestigatingtamingzero,ma2025generalreasoneradvancingllmreasoning,huang2025rzeroselfevolvingreasoningllm}, we report the mean@32 metric for the AIME24/25 benchmarks and greedy decoding accuracy (pass@1) for all the other mathematical datasets (See Appendix~\ref{appendix:eval}). 

\textbf{Models and Baselines.}
To demonstrate the scalability of \textbf{\sysname{}}, we conduct experiments on three base pretrained models: Qwen2.5-Math-1.5B, Qwen2.5-Math-7B~\cite{qwen2025qwen25technicalreport} and Qwen3-4B-Base~\cite{yang2025qwen3technicalreport}.
We compare with several representative baselines: (1)\textbf{ Pretrained model}, which evaluates all base pretrained models in a standard evaluation setting as static performance references. (2) \textbf{Self-Consistency}~\cite{wang2023selfconsistencyimproveschainthought}, which aggregates multiple reasoning paths via majority voting as a test-time scaling method. (3) \textbf{TTRL}~\cite{zuo2025ttrltesttimereinforcementlearning}, which adapts the model on unlabeled test instances using pseudo-labels. (4) \textbf{R-Zero}~\cite{huang2025rzeroselfevolvingreasoningllm}, which enables data-free optimization through Challenger–Solver co-evolution. Detailed descriptions of the baselines can be found in Appendix~\ref{baselines}.

\textbf{Implementation Details.}
Our framework is implemented based on the VeRL~\cite{Sheng_2025}. During synthesizer training, we set the rollout number for question generation to $M$$=$$4$. For each synthesized question, the solver sample $K$$=$$10$ responses to estimate the quality of each generated question with $\gamma$$=$$1.2$. To promote diversity, the coefficients for the similarity penalty terms are set to $\lambda_1$$=$$1.0$ and $\lambda_2$$=$$1.0$. 
For the solver training, we employ a rollout group size of $G$$=$$8$  and $\delta$$=$$0.25$ to explore reasoning paths effectively. The further detailed information can be found in Appendix~\ref{appendx:implementation-details}.

\subsection{Main Results}
\textbf{\sysname{} achieves superior performance across most tasks and models.} As shown in Table~\ref{tab:main_results}, \sysname{} consistently outperforms all baselines in average accuracy.  On the \text{Qwen2.5-Math-1.5B} model, \sysname{} elevates the average points from 17.30 to 41.49, yielding a massive improvement of \textbf{+24.19} points. This advantage extends effectively to larger model scales; on \text{Qwen2.5-Math-7B}, our method achieves an average points of 52.54, surpassing the standard test-time scaling baseline Self-Consistency by \textbf{+20.39} points. These results indicate that our active test-time training paradigm is significantly more effective than passive inference-time scaling, unlocking reasoning potential that standard consistency strategies cannot reach.

\textbf{\sysname{} outperforms existing self-evolving methods.} Compared to recent baselines such as R-Zero and TTRL, \sysname{} exhibits superior performance. While TTRL generally scores higher than R-Zero by optimizing on test questions, \sysname{} achieves further gains. For instance, on the \text{Qwen2.5-Math-7B} benchmark, our method surpasses TTRL by \textbf{+4.12} points on average ($48.42$$\to$$ 52.54$). On the \text{Qwen3-4B-Base} model, \sysname{} also maintains the lead with average 47.21 points, exceeding TTRL by \textbf{+3.62} points. This consistent improvement indicates that \sysname{} bridges the capability gap by introducing synthesized intermediate problems, whereas TTRL remains constrained by the fixed difficulty of raw test questions during test-time training.

\textbf{\sysname{} demonstrates substantial improvements on challenging benchmarks.} 
The performance difference between \sysname{} and other methods is particularly evident on difficult benchmarks such as AIME24/25. On AIME24 with the \text{Qwen2.5-Math-1.5B} model, TTRL achieves 13.23 points, whereas \sysname{} increases this to 19.79, achieving a substantial \textbf{+6.56} points. Similarly, on the \text{Qwen2.5-Math-7B} model, \sysname{} scores 19.90 points on AIME25, surpassing the TTRL's 14.06 points by \textbf{+5.84} points. This confirms that when TTRL struggles with unreliable pseudo-labels on intractable tasks, \sysname{} successfully extracts high-quality supervision from synthesized curriculum problems, enabling the model to learn even when the target questions are initially beyond its reach.

\begin{figure*}[t] 
    \centering
    
    \includegraphics[width=\textwidth]{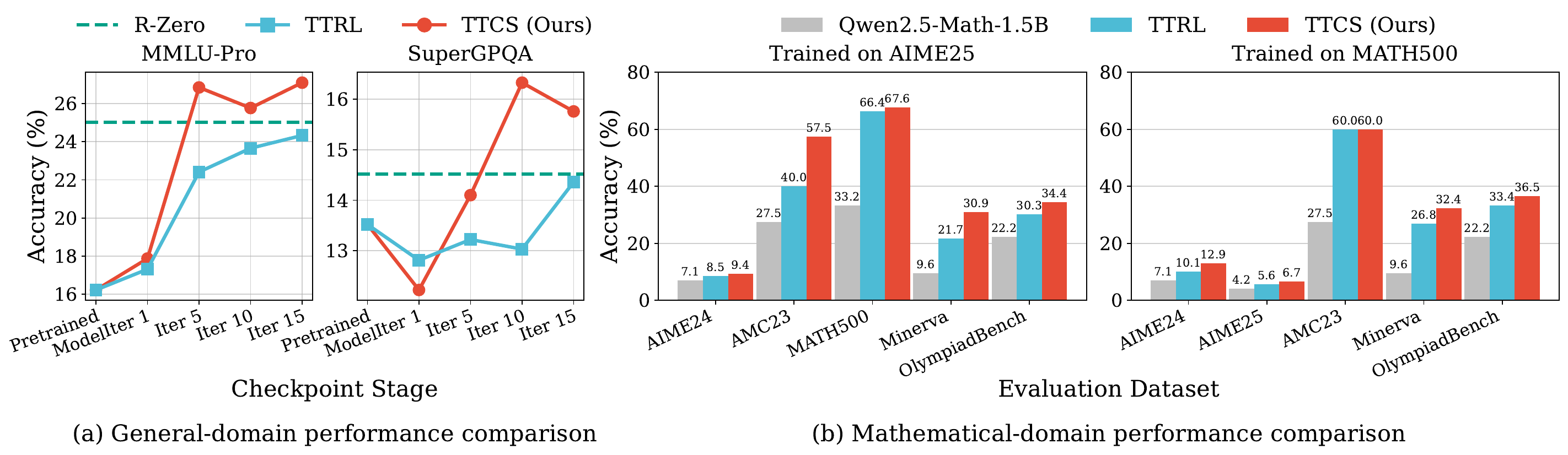} 
    \caption{Generalization analysis. 
\textbf{(a) General-domain generalization:} Accuracy trends of \sysname{} and TTRL on general-domain reasoning benchmarks (MMLU-Pro, SuperGPQA) during test-time training on AIME25. The green dashed line indicates the R-Zero baseline.
\textbf{(b) Mathematical-domain generalization:} Performance comparison on out-of-distribution (OOD) mathematical benchmarks.\label{fig:analysis}}
    
\end{figure*}

\subsection{Analysis}
In this section, we conduct comprehensive analysis experiments on \text{Qwen2.5-Math-1.5B} to obtain deeper insights about \textbf{\sysname{}}. We aim to answer the following research questions:

\textbf{Q1: Is the \sysname{} framework effective in enhancing reasoning capabilities in general domains beyond mathematics?}
To investigate the cross-domain generalization of \sysname{}, we conducted evaluations on challenging general-domain reasoning benchmarks, including MMLU-Pro~\cite{wang2024mmluprorobustchallengingmultitask} and SuperGPQA~\cite{pteam2025supergpqascalingllmevaluation}. 
Crucially, these evaluations are performed while the model undergoes online test-time self-evolving specifically on the AIME25 benchmark. Since R-Zero is unstable under extended iterations, we report its performance at the $2$-th iteration as a static baseline. As shown in Figure~\ref{fig:analysis}(a), \sysname{} demonstrates superior generalization capabilities. On MMLU-Pro, it surpasses the R-Zero baseline after 5 iterations and consistently outperforms TTRL. Similarly, on SuperGPQA, \sysname{} peaks significantly above R-Zero, whereas TTRL remains below the baseline. These results indicate that gains learned during mathematical self-evolution generalize to broader reasoning tasks. Additional results, including BBEH~\cite{kazemi2025bigbenchextrahard}, are provided in Appendix~\ref{appendix:genera-all}.

\textbf{Q2: Can the solver trained on a specific dataset generalize to unseen benchmarks?} To evaluate the out-of-domain generalization, we train the solver on a single test dataset and directly evaluate it on the other unseen benchmarks. Results are shown in Figure~\ref{fig:analysis}(b). The solver achieves consistent performance gains across all unseen datasets, indicating strong generalization. Specifically, even when trained solely on MATH-500, a dataset of moderate difficulty, the model still improves substantially on more challenging benchmarks, with accuracy on AIME24 increasing from 7.1 points to 12.9 points. This result suggests that during the co-evolving process of \sysname{}, the solver acquires universal mathematical reasoning logic. Rather than memorizing patterns from the test set, the model improves its ability to handle diverse and unseen mathematical problems. The results of other datasets are provided in Appendix~\ref{appendix:ood-all}.

\begin{table}[h]
\centering

\renewcommand{\arraystretch}{1.05}
\setlength{\tabcolsep}{2pt} 
\begin{tabular}{lccccc}
\toprule
\textbf{Method} & \textbf{AIME24} & \textbf{AIME25}  & \textbf{Minerva} & \textbf{AVG} \\
\midrule
Qwen2.5-Math-1.5B & 13.23  & 9.38 & 34.93 & 19.18 \\
\multicolumn{1}{l}{+Strong Synthesizer} & 16.35  & 10.21 & 38.97 & 21.84 \\
\rowcolor{rowblue} \multicolumn{1}{l}{+\textbf{TTCS(Ours)}} & \textbf{19.79}  & \textbf{13.33}  & \textbf{40.44} & \textbf{24.52}\\

\bottomrule
\end{tabular}
\caption{Performance comparison on the Qwen2.5-Math-1.5B model investigating the impact of synthesizer capability. The Strong synthesizer variant employs a fixed Qwen2.5-14B-Instruct model to generate questions.}
\label{tab:strong-syn}
\end{table}

\textbf{Q3: Is co-evolution more critical to \sysname{} than a stronger but static synthesizer?}
To further this, we replace the co-evolving 1.5B synthesizer with a frozen but significantly stronger \text{Qwen2.5-14B-Instruct}, while keeping the rest of the framework unchanged. As shown in Table~\ref{tab:strong-syn}, even with superior intrinsic capabilities, the static 14B synthesizer yields only a limited improvement of +2.66 points. In contrast, \sysname{} achieves a substantial gain of \textbf{+5.34} points, effectively doubling the performance boost of the stronger static baseline. This empirical evidence confirms that for test-time self-evolution, the adaptivity of the curriculum is far more decisive than the absolute strength of the teacher model.
\begin{wrapfigure}[12]{r}{0.5\textwidth} 
    \centering
    \includegraphics[width=0.48\textwidth]{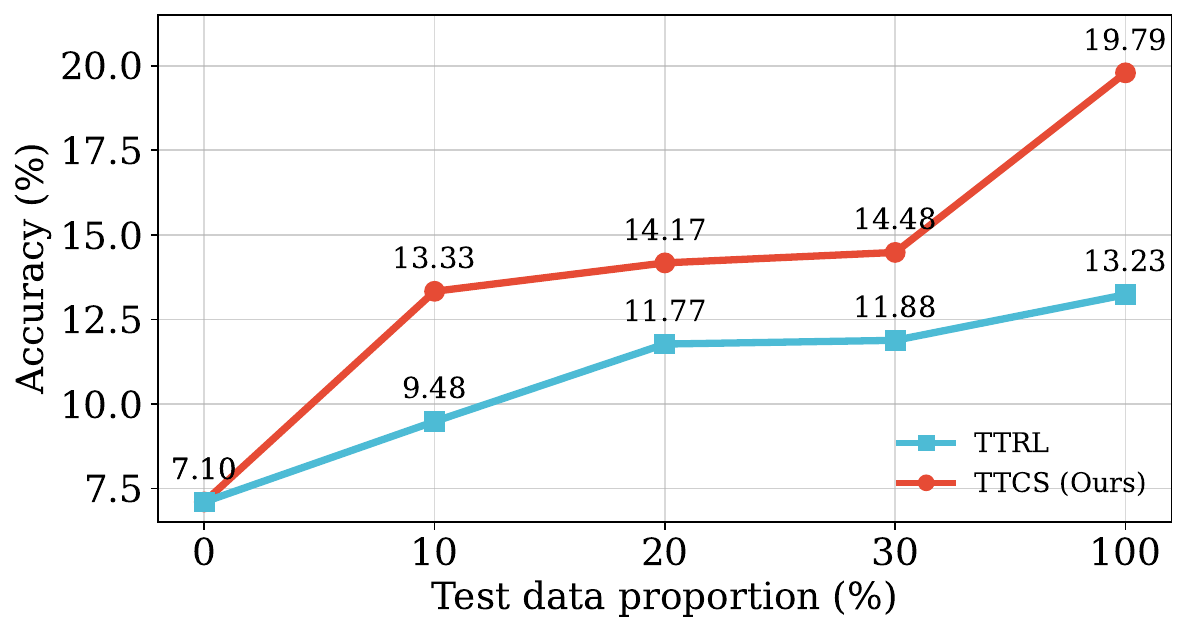} 
    \caption{Data efficiency analysis of test-time training on AIME24.}
    \label{fig:few-data}
\end{wrapfigure}
\textbf{Q4: Can \sysname{} remain effective under limited test data?}
To investigate efficacy in data-scarce scenarios, we evaluate our framework on AIME24 by restricting training data to varying proportions. As shown in Figure~\ref{fig:few-data}, starting from the pretrained model, \sysname{} consistently outperforms TTRL. Notably, with only 10\% data (3 questions), \sysname{} boosts accuracy to \textbf{13.33 points}, significantly surpassing TTRL's 9.48. This validates that \sysname{} effectively amplifies limited supervision via curriculum synthesis, ensuring stable self-evolution even with minimal test data.

\subsection{Ablation Study}
We conduct ablation studies on \text{Qwen2.5-Math-1.5B} to evaluate key components of \sysname{}, with results shown in Table~\ref{tab:ablation_study}.

\textbf{Without Synthesizer Training.} We freeze the synthesizer and use a static pretrained model to generate questions while allowing only the solver to evolve. Performance drops consistently across benchmarks, with accuracy on AMC23 decreasing from 62.50 points to 55.00 points. This shows that static question synthesis cannot adapt to the solver’s evolving capabilities, leading to questions that are either too easy or too difficult and thus provide ineffective learning signals.

\textbf{Without Online Data Filtering.} We remove the consistency-based online data filter and retain samples from all solver rollouts regardless of their difficulty. This ablation causes a clear performance decline across benchmarks, with Olympiad accuracy dropping from 36.05 points to 33.68 points. Without filtering at the capability boundary, the solver spends computation on low-value samples, which weakens the effective training signal.

\textbf{Without Diversity Penalties.} We remove the similarity penalties in the synthesizer's reward function that discourage copying from the reference test question ($\mathcal{R}_\text{ref}$) and redundancy within a generation group ($\mathcal{R}_\text{group}$), as described in Section~\ref{method:similarity_penalty}. Performance degrades under this setting, as the synthesizer tends to generate paraphrased or repetitive questions. The resulting lack of diversity can lead to overfitting or mode collapse and limit further improvement of the solver.

\begin{table}[!t]
\centering

\renewcommand{\arraystretch}{1.1}
\setlength{\tabcolsep}{2pt}

\begin{tabular}{l|cc|cc|cc}
\toprule
\textbf{Settings}
& \multicolumn{2}{c}{\textbf{AMC23}}
& \multicolumn{2}{c}{\textbf{Olympiad}}
& \multicolumn{2}{c}{\textbf{Minerva}} \\
\cmidrule(lr){2-3} \cmidrule(lr){4-5} \cmidrule(lr){6-7}
& \textbf{Score} & \textbf{$\Delta$}
& \textbf{Score} & \textbf{$\Delta$}
& \textbf{Score} & \textbf{$\Delta$} \\
\midrule
\textbf{TTCS (Full)}
& \textbf{62.50} & --
& \textbf{36.05} & --
& \textbf{40.44} & -- \\
\midrule
w/o Synthesizer Training
& 55.00 & \cellcolor{red!30}{$\downarrow$7.50}
& 32.79 & \cellcolor{red!20}{$\downarrow$3.26}
& 37.50 & \cellcolor{red!20}{$\downarrow$2.94} \\
w/o Online Data Filter
& 60.00 & \cellcolor{red!20}{$\downarrow$2.50}
& 33.68 & \cellcolor{red!20}{$\downarrow$2.37}
& 38.60 & \cellcolor{red!10}{$\downarrow$1.84} \\
w/o Diversity Penalty
& 55.00 & \cellcolor{red!30}{$\downarrow$7.50}
& 35.31 & \cellcolor{red!10}{$\downarrow$0.74}
& 39.34 & \cellcolor{red!10}{$\downarrow$1.10} \\
\bottomrule
\end{tabular}
\caption{\textbf{Ablation study} of TTCS components on the Qwen2.5-Math-1.5B model. Performance drops ($\Delta$) are computed relative to the full TTCS setting.}
\label{tab:ablation_study}
\end{table}

\section{Case Study}
\label{appendix:case_study}

\definecolor{evolve1}{RGB}{65, 105, 225}   
\definecolor{evolve2}{RGB}{220, 20, 60}    
\definecolor{evolve3}{RGB}{34, 139, 34}    
\definecolor{evolve4}{RGB}{148, 0, 211}    

To qualitatively demonstrate how our self-evolving framework progressively improves question synthesis quality, we present two representative cases from AIME24. For each case, we track the synthesized questions across training iterations (Iter 1, 5, 10, and 15), highlighting the evolution in problem structure and complexity.

We use color coding to indicate different types of evolution patterns observed in the synthesized questions:
\begin{itemize}
    \item \textcolor{evolve1}{\textbf{Blue (Structure Change)}}: Modifications to problem parameters, variables, or mathematical objects while preserving the core solution approach.
    \item \textcolor{evolve2}{\textbf{Red (Difficulty Increase)}}: Introduction of additional constraints, more complex conditions, or harder computational requirements.
    \item \textcolor{evolve3}{\textbf{Green (Domain Transfer)}}: Shift to different mathematical domains (e.g., from algebra to geometry, from discrete to continuous).
    \item \textcolor{evolve4}{\textbf{Purple (Complexity Growth)}}: Increase in structural complexity, such as nested functions, multi-step reasoning, or advanced mathematical concepts.
\end{itemize}

As shown in Table~\ref{tab:evolution_cases}, the model demonstrates increasing sophistication in question synthesis as training progresses. In early iterations, the synthesized questions tend to be simpler variations or direct modifications of the reference. By later iterations, the model generates questions that exhibit domain transfer, increased structural complexity, and novel problem formulations while maintaining mathematical validity.

\begin{table*}[t]
    \centering
    \small 
    \renewcommand{\arraystretch}{1.5} 
    
    \begin{tabularx}{\textwidth}{l X X}
        \toprule
        \textbf{Iteration} & \textbf{Case 1: Function Intersection} & \textbf{Case 2: Roots of Unity} \\
        \midrule
        
        \textbf{Test Question (AIME24)} & 
        Define $f(x)=|| x|-\tfrac{1}{2}|$ and $g(x)=|| x|-\tfrac{1}{4}|$. Find the number of intersections of the graphs of $y=4 g(f(\sin (2 \pi x)))$ and $x=4 g(f(\cos (3 \pi y)))$. &
        Let $\omega \neq 1$ be a 13th root of unity. Find the remainder when $\prod_{k=0}^{12}(2 - 2\omega^k + \omega^{2k})$ is divided by 1000. \\
        \midrule
        
        \textbf{Iter 1} & 
        Find the number of \textcolor{evolve3}{\textbf{integer solutions}} $(x,y)$ to the equation \textcolor{evolve1}{\textbf{$4x^2-16y^2+14x-24y+9=0$}}. &
        The polynomial \textcolor{evolve1}{\textbf{$f(x)=x^3+x^2-x-1$}} has three distinct roots, $p$, $q$, and $r$. What is $p^3q^3+r^3q^3+p^3r^3$? \\
        \midrule
        
        \textbf{Iter 5} & 
        How many points $(x,y)$ lie at intersections between the graphs of \textcolor{evolve2}{\textbf{$x=|\sin(\pi y)|$}} and \textcolor{evolve2}{\textbf{$y=|\sin(\pi x)|$}}? &
        Find the sum of all \textcolor{evolve3}{\textbf{complex numbers}} $w$ such that the real part of $w$ is greater than 0 and \textcolor{evolve2}{\textbf{$w^3 + (4 - \frac{58}{7}i)w^2 - (6 + 13i)w + 10 = 0$}}. \\
        \midrule
        
        \textbf{Iter 10} & 
        Find the number of points where the graphs of \textcolor{evolve4}{\textbf{$y=|x+1|+|x-1|$}} and \textcolor{evolve4}{\textbf{$y=|2x-1|+|3x-2|$}} intersect. &
        Let $\omega \neq 1$ be a 13th root of unity. Compute the remainder when \textcolor{evolve1}{\textbf{$\sum_{n=0}^{12} \omega^n$}} is divided by 1000. \\
        \midrule
        
        \textbf{Iter 15} & 
        Find the sum of the $x$-coordinates of the points where the graph of \textcolor{evolve4}{\textbf{$f(x)=\sin^{-1}\left(\frac{3x-x^3}{1+2x^2}\right)$}} intersects the line $y=\frac{\pi}{4}$, for \textcolor{evolve2}{\textbf{$-\sqrt{2} \leq x \leq \sqrt{2}$}}. &
        Let $N$ be the number of \textcolor{evolve3}{\textbf{ordered triples}} $(a, b, c)$ of positive integers such that \textcolor{evolve4}{\textbf{$a^2 + b^2 + c^2 = 2025$}}. Find $N$ modulo 1000. \\
        \bottomrule
    \end{tabularx}
     \caption{Evolution of synthesized questions across training iterations. Two representative cases from AIME24 are shown, demonstrating how our self-evolving framework progressively improves question synthesis quality.}
    \label{tab:evolution_cases}
\end{table*}

\section{Conclusion}
In this paper, we propose \textbf{\sysname{}}, a co-evolving test-time training framework for self-evolving. We address the critical failures of prior methods, specifically the limitation of unreliable pseudo-labels and the absence of intermediate difficulty learnable samples. By incorporating a capability-aware synthesizer, our approach dynamically constructs a curriculum of tractable variants to bridge the learning gap, converting noisy rewards into valid supervision. Empirical results confirm that \sysname{} delivers substantial gains on mathematical benchmarks and effectively generalizes to broader general-domain reasoning tasks. We view this work as a foundational step towards autonomous self-improvement and further plan to extend this framework to more useful and practical agentic applications.

\begin{ack}
We thank the reviewers for their insightful comments and constructive feedback. This work was supported by the National Natural Science Foundation of China (No. 62676342), the Open Competition for Innovative Projects of Xiamen (No. 3502Z20251012), the Natural Science Foundation of Fujian Province of China (No. 2024J011001), the Public Technology Service Platform Project of Xiamen (No. 3502Z20231043).
\end{ack}
\bibliographystyle{plain}
\bibliography{reference}

@misc{min2026orchestratingtokenssequencesdynamic,
      title={Orchestrating Tokens and Sequences: Dynamic Hybrid Policy Optimization for RLVR}, 
      author={Zijun Min and Bingshuai Liu and Ante Wang and Long Zhang and Anxiang Zeng and Haibo Zhang and Jinsong Su},
      year={2026},
      eprint={2601.05607},
      archivePrefix={arXiv},
      primaryClass={cs.LG},
      url={https://arxiv.org/abs/2601.05607}, 
}

@misc{liu2026healingentropycollapseenhancing,
      title={HEALing Entropy Collapse: Enhancing Exploration in Few-Shot RLVR via Hybrid-Domain Entropy Dynamics Alignment}, 
      author={Zhanyu Liu and Qingguo Hu and Ante Wang and Chenqing Liu and Zhishang Xiang and Hui Li and Delai Qiu and Jinsong Su},
      year={2026},
      eprint={2604.17928},
      archivePrefix={arXiv},
      primaryClass={cs.LG},
      url={https://arxiv.org/abs/2604.17928}, 
}

@misc{liu2026specrlacceleratingonpolicyreinforcement,
      title={SPEC-RL: Accelerating On-Policy Reinforcement Learning with Speculative Rollouts}, 
      author={Bingshuai Liu and Ante Wang and Zijun Min and Liang Yao and Haibo Zhang and Yang Liu and Xu Han and Peng Li and Anxiang Zeng and Jinsong Su},
      year={2026},
      eprint={2509.23232},
      archivePrefix={arXiv},
      primaryClass={cs.LG},
      url={https://arxiv.org/abs/2509.23232}, 
}

@misc{yu2025dapoopensourcellmreinforcement,
      title={DAPO: An Open-Source LLM Reinforcement Learning System at Scale}, 
      author={Qiying Yu and Zheng Zhang and Ruofei Zhu and Yufeng Yuan et al.},
      year={2025},
      eprint={2503.14476},
      archivePrefix={arXiv},
      primaryClass={cs.LG},
      url={https://arxiv.org/abs/2503.14476}, 
}

@misc{jin2025searchr1trainingllmsreason,
      title={Search-R1: Training LLMs to Reason and Leverage Search Engines with Reinforcement Learning}, 
      author={Bowen Jin and Hansi Zeng and Zhenrui Yue and Jinsung Yoon and Sercan Arik and Dong Wang and Hamed Zamani and Jiawei Han},
      year={2025},
      eprint={2503.09516},
      archivePrefix={arXiv},
      primaryClass={cs.CL},
      url={https://arxiv.org/abs/2503.09516}, 
}

@misc{shao2024deepseekmathpushinglimitsmathematical,
      title={DeepSeekMath: Pushing the Limits of Mathematical Reasoning in Open Language Models}, 
      author={Zhihong Shao and Peiyi Wang and Qihao Zhu and Runxin Xu and Junxiao Song and Xiao Bi and Haowei Zhang and Mingchuan Zhang and Y. K. Li and Y. Wu and Daya Guo},
      year={2024},
      eprint={2402.03300},
      archivePrefix={arXiv},
      primaryClass={cs.CL},
      url={https://arxiv.org/abs/2402.03300}, 
}

@misc{zuo2025ttrltesttimereinforcementlearning,
      title={TTRL: Test-Time Reinforcement Learning}, 
      author={Yuxin Zuo and Kaiyan Zhang and Li Sheng and Shang Qu and Ganqu Cui and Xuekai Zhu and Haozhan Li and Yuchen Zhang and Xinwei Long and Ermo Hua and Biqing Qi and Youbang Sun and Zhiyuan Ma and Lifan Yuan and Ning Ding and Bowen Zhou},
      year={2025},
      eprint={2504.16084},
      archivePrefix={arXiv},
      primaryClass={cs.CL},
      url={https://arxiv.org/abs/2504.16084}, 
}

@misc{huang2025rzeroselfevolvingreasoningllm,
      title={R-Zero: Self-Evolving Reasoning LLM from Zero Data}, 
      author={Chengsong Huang and Wenhao Yu and Xiaoyang Wang and Hongming Zhang and Zongxia Li and Ruosen Li and Jiaxin Huang and Haitao Mi and Dong Yu},
      year={2025},
      eprint={2508.05004},
      archivePrefix={arXiv},
      primaryClass={cs.LG},
      url={https://arxiv.org/abs/2508.05004}, 
}

@misc{wei2025sftsecondrlupt,
      title={First SFT, Second RL, Third UPT: Continual Improving Multi-Modal LLM Reasoning via Unsupervised Post-Training}, 
      author={Lai Wei and Yuting Li and Chen Wang and Yue Wang and Linghe Kong and Weiran Huang and Lichao Sun},
      year={2025},
      eprint={2505.22453},
      archivePrefix={arXiv},
      primaryClass={cs.CL},
      url={https://arxiv.org/abs/2505.22453}, 
}

@misc{fang2025serlselfplayreinforcementlearning,
      title={SeRL: Self-Play Reinforcement Learning for Large Language Models with Limited Data}, 
      author={Wenkai Fang and Shunyu Liu and Yang Zhou and Kongcheng Zhang and Tongya Zheng and Kaixuan Chen and Mingli Song and Dacheng Tao},
      year={2025},
      eprint={2505.20347},
      archivePrefix={arXiv},
      primaryClass={cs.CL},
      url={https://arxiv.org/abs/2505.20347}, 
}

@misc{chen2025spcevolvingselfplaycritic,
      title={SPC: Evolving Self-Play Critic via Adversarial Games for LLM Reasoning}, 
      author={Jiaqi Chen and Bang Zhang and Ruotian Ma and Peisong Wang and Xiaodan Liang and Zhaopeng Tu and Xiaolong Li and Kwan-Yee K. Wong},
      year={2025},
      eprint={2504.19162},
      archivePrefix={arXiv},
      primaryClass={cs.CL},
      url={https://arxiv.org/abs/2504.19162}, 
}

@misc{zhao2025majorityrightrltraining,
      title={The Majority is not always right: RL training for solution aggregation}, 
      author={Wenting Zhao and Pranjal Aggarwal and Swarnadeep Saha and Asli Celikyilmaz and Jason Weston and Ilia Kulikov},
      year={2025},
      eprint={2509.06870},
      archivePrefix={arXiv},
      primaryClass={cs.CL},
      url={https://arxiv.org/abs/2509.06870}, 
}

@misc{deepseekai2025deepseekr1incentivizingreasoningcapability,
      title={DeepSeek-R1: Incentivizing Reasoning Capability in LLMs via Reinforcement Learning}, 
      author={DeepSeek-AI and Daya Guo et al.},
      year={2025},
      eprint={2501.12948},
      archivePrefix={arXiv},
      primaryClass={cs.CL},
      url={https://arxiv.org/abs/2501.12948}, 
}

@misc{kimiteam2025kimik2openagentic,
      title={Kimi K2: Open Agentic Intelligence}, 
      author={Kimi Team and Yifan Bai and Yiping Bao et al.},
      year={2025},
      eprint={2507.20534},
      archivePrefix={arXiv},
      primaryClass={cs.LG},
      url={https://arxiv.org/abs/2507.20534}, 
}

@misc{yang2025qwen3technicalreport,
      title={Qwen3 Technical Report}, 
      author={An Yang and Anfeng Li and Baosong Yang et al.},
      year={2025},
      eprint={2505.09388},
      archivePrefix={arXiv},
      primaryClass={cs.CL},
      url={https://arxiv.org/abs/2505.09388}, 
}

@misc{grattafiori2024llama3herdmodels,
      title={The Llama 3 Herd of Models}, 
      author={Aaron Grattafiori and Abhimanyu Dubey and Abhinav Jauhri et al.},
      year={2024},
      eprint={2407.21783},
      archivePrefix={arXiv},
      primaryClass={cs.AI},
      url={https://arxiv.org/abs/2407.21783}, 
}

@misc{qwen2025qwen25technicalreport,
      title={Qwen2.5 Technical Report}, 
      author={Qwen and : and An Yang and Baosong Yang and Beichen Zhang and Binyuan Hui and Bo Zheng and Bowen Yu and Chengyuan Li and Dayiheng Liu and Fei Huang and Haoran Wei and Huan Lin and Jian Yang and Jianhong Tu and Jianwei Zhang and Jianxin Yang and Jiaxi Yang and Jingren Zhou and Junyang Lin and Kai Dang and Keming Lu and Keqin Bao and Kexin Yang and Le Yu and Mei Li and Mingfeng Xue and Pei Zhang and Qin Zhu and Rui Men and Runji Lin and Tianhao Li and Tianyi Tang and Tingyu Xia and Xingzhang Ren and Xuancheng Ren and Yang Fan and Yang Su and Yichang Zhang and Yu Wan and Yuqiong Liu and Zeyu Cui and Zhenru Zhang and Zihan Qiu},
      year={2025},
      eprint={2412.15115},
      archivePrefix={arXiv},
      primaryClass={cs.CL},
      url={https://arxiv.org/abs/2412.15115}, 
}

@misc{shafayat2025largereasoningmodelsselftrain,
      title={Can Large Reasoning Models Self-Train?}, 
      author={Sheikh Shafayat and Fahim Tajwar and Ruslan Salakhutdinov and Jeff Schneider and Andrea Zanette},
      year={2025},
      eprint={2505.21444},
      archivePrefix={arXiv},
      primaryClass={cs.LG},
      url={https://arxiv.org/abs/2505.21444}, 
}

@misc{ma2025generalreasoneradvancingllmreasoning,
      title={General-Reasoner: Advancing LLM Reasoning Across All Domains}, 
      author={Xueguang Ma and Qian Liu and Dongfu Jiang and Ge Zhang and Zejun Ma and Wenhu Chen},
      year={2025},
      eprint={2505.14652},
      archivePrefix={arXiv},
      primaryClass={cs.CL},
      url={https://arxiv.org/abs/2505.14652}, 
}

@misc{lin2025learningsolveverifyselfplay,
      title={Learning to Solve and Verify: A Self-Play Framework for Code and Test Generation}, 
      author={Zi Lin and Sheng Shen and Jingbo Shang and Jason Weston and Yixin Nie},
      year={2025},
      eprint={2502.14948},
      archivePrefix={arXiv},
      primaryClass={cs.SE},
      url={https://arxiv.org/abs/2502.14948}, 
}

@misc{zeng2025simplerlzooinvestigatingtamingzero,
      title={SimpleRL-Zoo: Investigating and Taming Zero Reinforcement Learning for Open Base Models in the Wild}, 
      author={Weihao Zeng and Yuzhen Huang and Qian Liu and Wei Liu and Keqing He and Zejun Ma and Junxian He},
      year={2025},
      eprint={2503.18892},
      archivePrefix={arXiv},
      primaryClass={cs.LG},
      url={https://arxiv.org/abs/2503.18892}, 
}

@misc{pteam2025supergpqascalingllmevaluation,
      title={SuperGPQA: Scaling LLM Evaluation across 285 Graduate Disciplines}, 
      author={P Team and Xinrun Du and Yifan Yao and Kaijing Ma et al.},
      year={2025},
      eprint={2502.14739},
      archivePrefix={arXiv},
      primaryClass={cs.CL},
      url={https://arxiv.org/abs/2502.14739}, 
}

@misc{briesch2024largelanguagemodelssuffer,
      title={Large Language Models Suffer From Their Own Output: An Analysis of the Self-Consuming Training Loop}, 
      author={Martin Briesch and Dominik Sobania and Franz Rothlauf},
      year={2024},
      eprint={2311.16822},
      archivePrefix={arXiv},
      primaryClass={cs.LG},
      url={https://arxiv.org/abs/2311.16822}, 
}

@misc{liang2025swsselfawareweaknessdrivenproblem,
      title={SwS: Self-aware Weakness-driven Problem Synthesis in Reinforcement Learning for LLM Reasoning}, 
      author={Xiao Liang and Zhong-Zhi Li and Yeyun Gong and Yang Wang and Hengyuan Zhang and Yelong Shen and Ying Nian Wu and Weizhu Chen},
      year={2025},
      eprint={2506.08989},
      archivePrefix={arXiv},
      primaryClass={cs.LG},
      url={https://arxiv.org/abs/2506.08989}, 
}

@misc{zhou2025selfchallenginglanguagemodelagents,
      title={Self-Challenging Language Model Agents}, 
      author={Yifei Zhou and Sergey Levine and Jason Weston and Xian Li and Sainbayar Sukhbaatar},
      year={2025},
      eprint={2506.01716},
      archivePrefix={arXiv},
      primaryClass={cs.AI},
      url={https://arxiv.org/abs/2506.01716}, 
}

@inproceedings{zhang2025weaving,
  title={Weaving context across images: Improving vision-language models through focus-centric visual chains},
  author={Zhang, Juntian and Cheng, Chuanqi and Liu, Yuhan and Liu, Wei and Luan, Jian and Yan, Rui},
  booktitle={Proceedings of the 63rd Annual Meeting of the Association for Computational Linguistics (Volume 1: Long Papers)},
  pages={27782--27798},
  year={2025}
}

@article{zhang2025viper,
  title={Viper: Empowering the self-evolution of visual perception abilities in vision-language model},
  author={Zhang, Juntian and Jin, Song and Cheng, Chuanqi and Liu, Yuhan and Lin, Yankai and Zhang, Xun and Zhang, Yufei and Jiang, Fei and Yin, Guojun and Lin, Wei and others},
  journal={arXiv preprint arXiv:2510.24285},
  year={2025}
}

@misc{wang2023selfconsistencyimproveschainthought,
      title={Self-Consistency Improves Chain of Thought Reasoning in Language Models}, 
      author={Xuezhi Wang and Jason Wei and Dale Schuurmans and Quoc Le and Ed Chi and Sharan Narang and Aakanksha Chowdhery and Denny Zhou},
      year={2023},
      eprint={2203.11171},
      archivePrefix={arXiv},
      primaryClass={cs.CL},
      url={https://arxiv.org/abs/2203.11171}, 
}

@misc{chen2024selfplayfinetuningconvertsweak,
      title={Self-Play Fine-Tuning Converts Weak Language Models to Strong Language Models}, 
      author={Zixiang Chen and Yihe Deng and Huizhuo Yuan and Kaixuan Ji and Quanquan Gu},
      year={2024},
      eprint={2401.01335},
      archivePrefix={arXiv},
      primaryClass={cs.LG},
      url={https://arxiv.org/abs/2401.01335}, 
}

@misc{zhao2025absolutezeroreinforcedselfplay,
      title={Absolute Zero: Reinforced Self-play Reasoning with Zero Data}, 
      author={Andrew Zhao and Yiran Wu and Yang Yue and Tong Wu and Quentin Xu and Yang Yue and Matthieu Lin and Shenzhi Wang and Qingyun Wu and Zilong Zheng and Gao Huang},
      year={2025},
      eprint={2505.03335},
      archivePrefix={arXiv},
      primaryClass={cs.LG},
      url={https://arxiv.org/abs/2505.03335}, 
}

@misc{liu2025spiceselfplaycorpusenvironments,
      title={SPICE: Self-Play In Corpus Environments Improves Reasoning}, 
      author={Bo Liu and Chuanyang Jin and Seungone Kim and Weizhe Yuan and Wenting Zhao and Ilia Kulikov and Xian Li and Sainbayar Sukhbaatar and Jack Lanchantin and Jason Weston},
      year={2025},
      eprint={2510.24684},
      archivePrefix={arXiv},
      primaryClass={cs.CL},
      url={https://arxiv.org/abs/2510.24684}, 
}

@misc{akyürek2025surprisingeffectivenesstesttimetraining,
      title={The Surprising Effectiveness of Test-Time Training for Few-Shot Learning}, 
      author={Ekin Akyürek and Mehul Damani and Adam Zweiger and Linlu Qiu and Han Guo and Jyothish Pari and Yoon Kim and Jacob Andreas},
      year={2025},
      eprint={2411.07279},
      archivePrefix={arXiv},
      primaryClass={cs.AI},
      url={https://arxiv.org/abs/2411.07279}, 
}

@misc{snell2024scalingllmtesttimecompute,
      title={Scaling LLM Test-Time Compute Optimally can be More Effective than Scaling Model Parameters}, 
      author={Charlie Snell and Jaehoon Lee and Kelvin Xu and Aviral Kumar},
      year={2024},
      eprint={2408.03314},
      archivePrefix={arXiv},
      primaryClass={cs.LG},
      url={https://arxiv.org/abs/2408.03314}, 
}

@misc{liu20251bllmsurpass405b,
      title={Can 1B LLM Surpass 405B LLM? Rethinking Compute-Optimal Test-Time Scaling}, 
      author={Runze Liu and Junqi Gao and Jian Zhao and Kaiyan Zhang and Xiu Li and Biqing Qi and Wanli Ouyang and Bowen Zhou},
      year={2025},
      eprint={2502.06703},
      archivePrefix={arXiv},
      primaryClass={cs.CL},
      url={https://arxiv.org/abs/2502.06703}, 
}

@article{hubert2025olympiad,
  title={Olympiad-level formal mathematical reasoning with reinforcement learning},
  author={Hubert, Thomas and Mehta, Rishi and Sartran, Laurent and Horv{\'a}th, Mikl{\'o}s Z and {\v{Z}}u{\v{z}}i{\'c}, Goran and Wieser, Eric and Huang, Aja and Schrittwieser, Julian and Schroecker, Yannick and Masoom, Hussain and others},
  journal={Nature},
  pages={1--3},
  year={2025},
  publisher={Nature Publishing Group UK London}
}

@misc{MATH-500,
      title={Measuring Mathematical Problem Solving With the MATH Dataset}, 
      author={Dan Hendrycks and Collin Burns and Saurav Kadavath and Akul Arora and Steven Basart and Eric Tang and Dawn Song and Jacob Steinhardt},
      year={2021},
      eprint={2103.03874},
      archivePrefix={arXiv},
      primaryClass={cs.LG},
      url={https://arxiv.org/abs/2103.03874}, 
}

@misc{Minerva,
      title={Solving Quantitative Reasoning Problems with Language Models}, 
      author={Aitor Lewkowycz and Anders Andreassen and David Dohan and Ethan Dyer and Henryk Michalewski and Vinay Ramasesh and Ambrose Slone and Cem Anil and Imanol Schlag and Theo Gutman-Solo and Yuhuai Wu and Behnam Neyshabur and Guy Gur-Ari and Vedant Misra},
      year={2022},
      eprint={2206.14858},
      archivePrefix={arXiv},
      primaryClass={cs.CL},
      url={https://arxiv.org/abs/2206.14858}, 
}

@misc{OlympiadBench,
      title={OlympiadBench: A Challenging Benchmark for Promoting AGI with Olympiad-Level Bilingual Multimodal Scientific Problems}, 
      author={Chaoqun He and Renjie Luo and Yuzhuo Bai and Shengding Hu and Zhen Leng Thai and Junhao Shen and Jinyi Hu and Xu Han and Yujie Huang and Yuxiang Zhang and Jie Liu and Lei Qi and Zhiyuan Liu and Maosong Sun},
      year={2024},
      eprint={2402.14008},
      archivePrefix={arXiv},
      primaryClass={cs.CL},
      url={https://arxiv.org/abs/2402.14008}, 
}

@inproceedings{Sheng_2025, series={EuroSys ’25},
   title={HybridFlow: A Flexible and Efficient RLHF Framework},
   url={http://dx.doi.org/10.1145/3689031.3696075},
   DOI={10.1145/3689031.3696075},
   booktitle={Proceedings of the Twentieth European Conference on Computer Systems},
   publisher={ACM},
   author={Sheng, Guangming and Zhang, Chi and Ye, Zilingfeng and Wu, Xibin and Zhang, Wang and Zhang, Ru and Peng, Yanghua and Lin, Haibin and Wu, Chuan},
   year={2025},
   month=mar, pages={1279–1297},
   collection={EuroSys ’25}
}

@article{shumailov2024ai,
  title={AI models collapse when trained on recursively generated data},
  author={Shumailov, Ilia and Shumaylov, Zakhar and Zhao, Yiren and Papernot, Nicolas and Anderson, Ross and Gal, Yarin},
  journal={Nature},
  volume={631},
  number={8022},
  pages={755--759},
  year={2024},
  publisher={Nature Publishing Group UK London}
}

@misc{huang2022largelanguagemodelsselfimprove,
      title={Large Language Models Can Self-Improve}, 
      author={Jiaxin Huang and Shixiang Shane Gu and Le Hou and Yuexin Wu and Xuezhi Wang and Hongkun Yu and Jiawei Han},
      year={2022},
      eprint={2210.11610},
      archivePrefix={arXiv},
      primaryClass={cs.CL},
      url={https://arxiv.org/abs/2210.11610}, 
}

@article{silver2025welcome,
  title={Welcome to the era of experience},
  author={Silver, David and Sutton, Richard S},
  journal={Google AI},
  volume={1},
  year={2025}
}

@article{409cf137-dbb5-3eb1-8cfe-0743c3dc925f,
 ISSN = {01621459, 1537274X},
 URL = {http://www.jstor.org/stable/2282952},
 author = {Wassily Hoeffding},
 journal = {Journal of the American Statistical Association},
 number = {301},
 pages = {13--30},
 publisher = {[American Statistical Association, Taylor & Francis, Ltd.]},
 title = {Probability Inequalities for Sums of Bounded Random Variables},
 urldate = {2026-01-14},
 volume = {58},
 year = {1963}
}

@misc{sun2020testtimetrainingselfsupervisiongeneralization,
      title={Test-Time Training with Self-Supervision for Generalization under Distribution Shifts}, 
      author={Yu Sun and Xiaolong Wang and Zhuang Liu and John Miller and Alexei A. Efros and Moritz Hardt},
      year={2020},
      eprint={1909.13231},
      archivePrefix={arXiv},
      primaryClass={cs.LG},
      url={https://arxiv.org/abs/1909.13231}, 
}

@misc{kazemi2025bigbenchextrahard,
      title={BIG-Bench Extra Hard}, 
      author={Mehran Kazemi and Bahare Fatemi and Hritik Bansal and John Palowitch and Chrysovalantis Anastasiou and Sanket Vaibhav Mehta and Lalit K. Jain and Virginia Aglietti and Disha Jindal and Peter Chen and Nishanth Dikkala and Gladys Tyen and Xin Liu and Uri Shalit and Silvia Chiappa and Kate Olszewska and Yi Tay and Vinh Q. Tran and Quoc V. Le and Orhan Firat},
      year={2025},
      eprint={2502.19187},
      archivePrefix={arXiv},
      primaryClass={cs.CL},
      url={https://arxiv.org/abs/2502.19187}, 
}

@misc{wang2024mmluprorobustchallengingmultitask,
      title={MMLU-Pro: A More Robust and Challenging Multi-Task Language Understanding Benchmark}, 
      author={Yubo Wang and Xueguang Ma and Ge Zhang and Yuansheng Ni and Abhranil Chandra and Shiguang Guo and Weiming Ren and Aaran Arulraj and Xuan He and Ziyan Jiang and Tianle Li and Max Ku and Kai Wang and Alex Zhuang and Rongqi Fan and Xiang Yue and Wenhu Chen},
      year={2024},
      eprint={2406.01574},
      archivePrefix={arXiv},
      primaryClass={cs.CL},
      url={https://arxiv.org/abs/2406.01574}, 
}

@misc{comanici2025gemini25pushingfrontier,
      title={Gemini 2.5: Pushing the Frontier with Advanced Reasoning, Multimodality, Long Context, and Next Generation Agentic Capabilities}, 
      author={Gheorghe Comanici and Eric Bieber and Mike Schaekermann et al.},
      year={2025},
      eprint={2507.06261},
      archivePrefix={arXiv},
      primaryClass={cs.CL},
      url={https://arxiv.org/abs/2507.06261}, 
}

@misc{selfinstruct,
  title={Self-Instruct: Aligning Language Model with Self Generated Instructions},
  author={Wang, Yizhong and Kordi, Yeganeh and Mishra, Swaroop and Liu, Alisa and Smith, Noah A. and Khashabi, Daniel and Hajishirzi, Hannaneh},
  journal={arXiv preprint arXiv:2212.10560},
  year={2022}
}

@misc{geminiteam2024gemini15unlockingmultimodal,
      title={Gemini 1.5: Unlocking multimodal understanding across millions of tokens of context}, 
      author={Gemini Team and Petko Georgiev and Ving Ian Lei and Ryan Burnell et al.},
      year={2024},
      eprint={2403.05530},
      archivePrefix={arXiv},
      primaryClass={cs.CL},
      url={https://arxiv.org/abs/2403.05530}, 
}

@misc{li2025limrrlscaling,
      title={LIMR: Less is More for RL Scaling}, 
      author={Xuefeng Li and Haoyang Zou and Pengfei Liu},
      year={2025},
      eprint={2502.11886},
      archivePrefix={arXiv},
      primaryClass={cs.LG},
      url={https://arxiv.org/abs/2502.11886}, 
}

@article{he2025visplay,
  title={Visplay: Self-evolving vision-language models from images},
  author={He, Yicheng and Huang, Chengsong and Li, Zongxia and Huang, Jiaxin and Yang, Yonghui},
  journal={arXiv preprint arXiv:2511.15661},
  year={2025}
}

@article{yu2025guided,
  title={Guided self-evolving llms with minimal human supervision},
  author={Yu, Wenhao and Liang, Zhenwen and Huang, Chengsong and Panaganti, Kishan and Fang, Tianqing and Mi, Haitao and Yu, Dong},
  journal={arXiv preprint arXiv:2512.02472},
  year={2025}
}


\appendix

\onecolumn      

\section{Experimental Settings}

In this section, we provide detailed descriptions of experimental settings used in \textbf{\sysname{}}. 
\subsection{Datasets}\label{appendix:datasets}

\paragraph{Competition-Level Mathematics.} To assess advanced reasoning capabilities, we utilize the following rigorous testbeds:
\begin{itemize}
    \item \textbf{AMC23}~\cite{zeng2025simplerlzooinvestigatingtamingzero}: A series of examinations used to identify and foster mathematical talent.
    \item \textbf{AIME24\&25}~\cite{li2025limrrlscaling}: We employ the 2024 and 2025 versions of the AIME. These problems serve as a proxy for competition-level difficulty, requiring multi-step reasoning and deep mathematical insight.
\end{itemize}

\paragraph{Standard Mathematical Benchmarks.} We include a set of widely used datasets to evaluate fundamental proficiency:
\begin{itemize}
    \item \textbf{MATH-500}~\cite{MATH-500}: A subset of the MATH dataset designed for efficient evaluation of mathematical problem-solving skills.
    \item \textbf{Minerva}~\cite{Minerva}: A collection of STEM problems covering a wide range of difficulty levels.
    \item \textbf{OlympiadBench}~\cite{OlympiadBench}: A comprehensive benchmark featuring Olympiad-level problems from various competitions.
\end{itemize}

\paragraph{General-Domain Benchmarks.} To evaluate the generalization ability of our framework beyond pure mathematics, we extend our analysis to:
\begin{itemize}
    \item \textbf{BBEH}~\cite{kazemi2025bigbenchextrahard}: Big Bench Extra Hard, focusing on tasks where language models traditionally struggle.
    \item \textbf{MMLU-Pro}~\cite{wang2024mmluprorobustchallengingmultitask}: A robust and challenging multi-task benchmark designed to push the limits of language understanding and reasoning.
    \item \textbf{SuperGPQA}~\cite{pteam2025supergpqascalingllmevaluation}: A dataset aimed at scaling LLM evaluation with graduate-level questions.
\end{itemize}

\subsection{Evaluation Metrics}\label{appendix:eval}

Consistent with prior works~\cite{zeng2025simplerlzooinvestigatingtamingzero,ma2025generalreasoneradvancingllmreasoning,huang2025rzeroselfevolvingreasoningllm}, we employ specific metrics tailored to the nature of each benchmark. For mathematical problems, to account for potential format variations in model outputs, we additionally use GPT-4o-mini to assist in judging whether the predicted answer matches the ground truth.

\begin{itemize}
    \item \textbf{Mean@32:} For the highly challenging AIME24 and AIME25 benchmarks, single-run performance can be unstable. To provide a robust estimate of the model's capability, we generate 32 solutions for each question using stochastic sampling (temperature $T=0.6$). We verify the correctness of each sample and report the average accuracy across all 32 samples, averaged over the entire dataset. This serves as an unbiased approximation of the model's expected pass rate.
    \item \textbf{Greedy Decoding (Pass@1):} For AMC, MATH-500, Minerva, and OlympiadBench, we strictly adhere to the standard evaluation setting. We generate a single solution using greedy decoding (temperature $T=0$) to evaluate the model's most confident reasoning trajectory. The final answer is extracted and compared with the ground truth.
    \item \textbf{Exact Match (EM):} For the general-domain benchmarks BBEH, MMLU-Pro, and SuperGPQA, we report Exact Match accuracy via greedy decoding (temperature $T=0.0$). We extract the predicted option or key phrase and check for a strict match against the ground truth label.
\end{itemize}
 
\subsection{Baselines}
\label{baselines}
We compare \textbf{\sysname{}} against these baselines:

\begin{itemize}
    \item \textbf{Pretrained Model:} We evaluate the diverse set of backbones introduced above in a standard zero-shot setting. This serves as a static performance benchmark to quantify the gains achieved by our method.

    \item \textbf{Self-Consistency} ~\cite{wang2023selfconsistencyimproveschainthought}: To account for reasoning stability, we employ Self-Consistency as a test-time scaling baseline. It enhances reasoning reliability by sampling multiple reasoning paths and aggregating the final answers via majority voting.

    \item \textbf{Test-Time Reinforcement Learning (TTRL)}~\cite{zuo2025ttrltesttimereinforcementlearning}: We adopt TTRL as a representative baseline for test-time training. This approach conducts reinforcement learning directly on unlabeled test instances by utilizing repeated sampling rollouts to estimate pseudo-labels via majority voting.

    \item \textbf{R-Zero}~\cite{huang2025rzeroselfevolvingreasoningllm}: We also compare against R-Zero, a method enabling fully data-free Challenger-Solver co-evolution. It allows the model to self-evolve its reasoning capabilities through reinforced self-play without relying on ground-truth annotations.
\end{itemize}

\subsection{Implementation Details}\label{appendx:implementation-details}
In this section, we elaborate on the detailed experimental settings and hyperparameter configurations used to train the \sysname{} framework. Table~\ref{tab:experiment_details} summarizes the standard training parameters for both the synthesizer and solver agents. Notably, for challenging benchmarks such as \textbf{AIME24}, \textbf{AIME25}, and \textbf{AMC23}, we specifically adjust the batch size to 16 and the solver's rollout group size to 16 to ensure training stability and efficient resource utilization.
\begin{table}[h]
    \centering
    
    \resizebox{0.5\linewidth}{!}{ 
    \begin{tabular}{lcc}
        \toprule
        \textbf{Config} &\multicolumn{1}{c}{\textbf{Synthesizer Training}} & \multicolumn{1}{c}{\textbf{Solver Training}} \\
        \midrule
        Batch Size & 32 & 64 \\
        Learning Rate & 1e-6  & 1e-6 \\
        Weight Decay &0.01 &0.01\\ 
        KL Coefficient & 0.01  & 0.01 \\
        Max Steps & 5 & 15 \\
        Rollout Group Size & 4 & 8 \\
        Rollout Temperature & 1.0 & 1.0 \\
        Rollout Top-p & 0.95 & 0.95 \\
        \bottomrule
    \end{tabular}
    }
    \caption{Detailed training configurations and hyper-parameters for the synthesizer and solver agents.}
    \label{tab:experiment_details}
\end{table}

\section{Additional Experimental Reults}
In this appendix, we provide a comprehensive suite of additional experimental results to further validate the robustness and generalization capabilities of \sysname{}. Specifically, we present the full spectrum of cross-task transfer performance across various mathematical benchmarks and extend our evaluation to general-domain reasoning tasks, demonstrating that the reasoning capability acquired by our method are not limited to specific datasets.

\subsection{General-Domain Performance}\label{appendix:genera-all}
\begin{figure*}[t] 
    \centering
    
    \includegraphics[width=\textwidth]{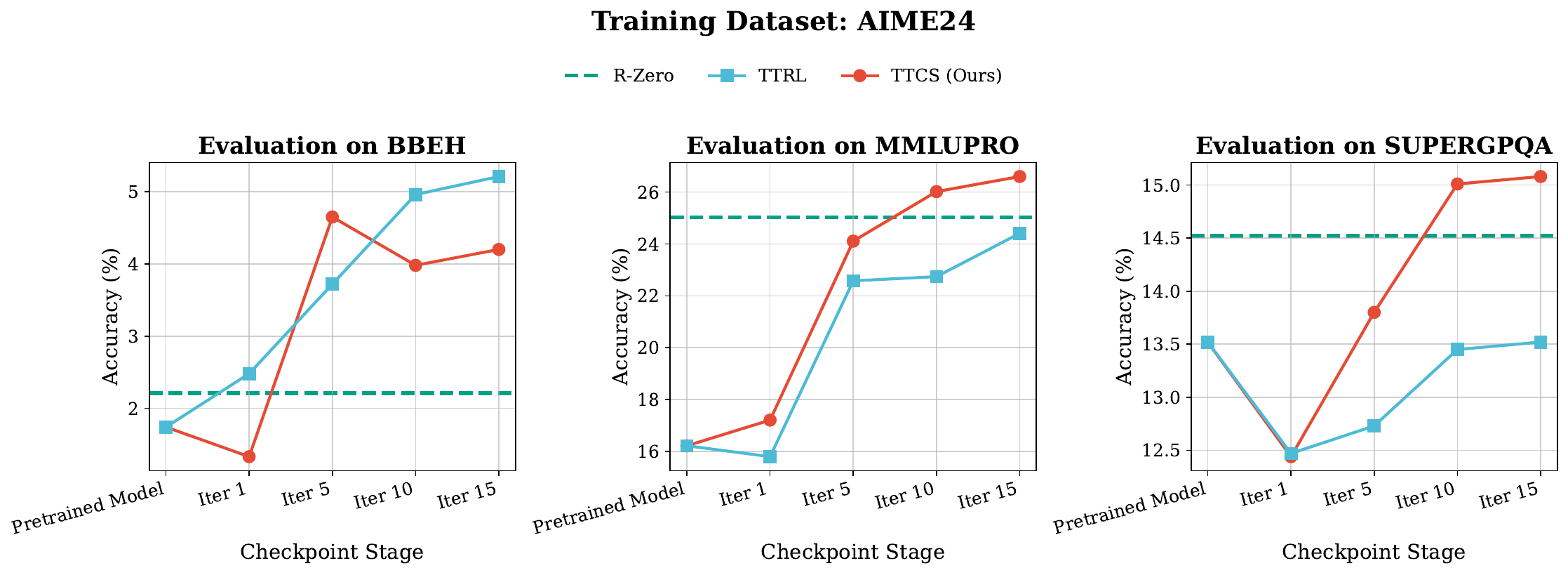} 
    \caption{The general-domain performance comparison of  \textbf{TTCS} and the other baselines when TTCS and TTRL trained on AIME24 dataset.}
    \label{fig:appendix-general-AIME24}
\end{figure*}

\begin{figure*}[t] 
    \centering
    
    \includegraphics[width=\textwidth]{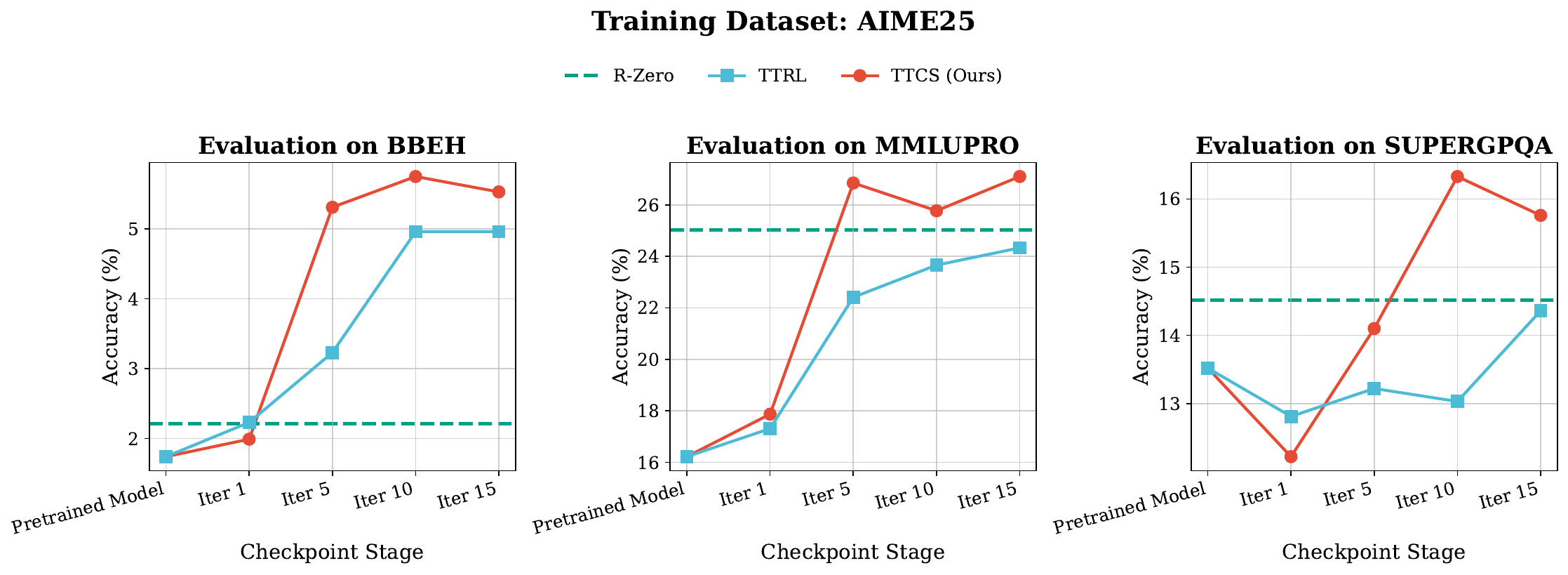} 
    \caption{The general-domain performance comparison of  \textbf{TTCS} and the other baselines when TTCS and TTRL trained on AIME25 dataset.}
    \label{fig:appendix-general-AIME25}
\end{figure*}

We further investigate whether the logical capabilities improved through mathematical reasoning can transfer to broader general domains. Figure~\ref{fig:appendix-general-AIME24} and Figure~\ref{fig:appendix-general-AIME25} illustrate the generalization performance on general reasoning benchmarks (e.g.,BBEH, MMLU-Pro and SuperGPQA) when the model is trained on \textbf{AIME24} and \textbf{AIME25}, respectively. In both scenarios, \sysname{} exhibits a substantial advantage over the baselines, suggesting that the high-quality reasoning chains synthesized during our training process 
facilitate a positive transfer to non-mathematical complex reasoning tasks.

\subsection{Out-of-Domain Mathematical Performance}\label{appendix:ood-all}
\begin{figure*}[t] 
    \centering
    
    \includegraphics[width=\textwidth]{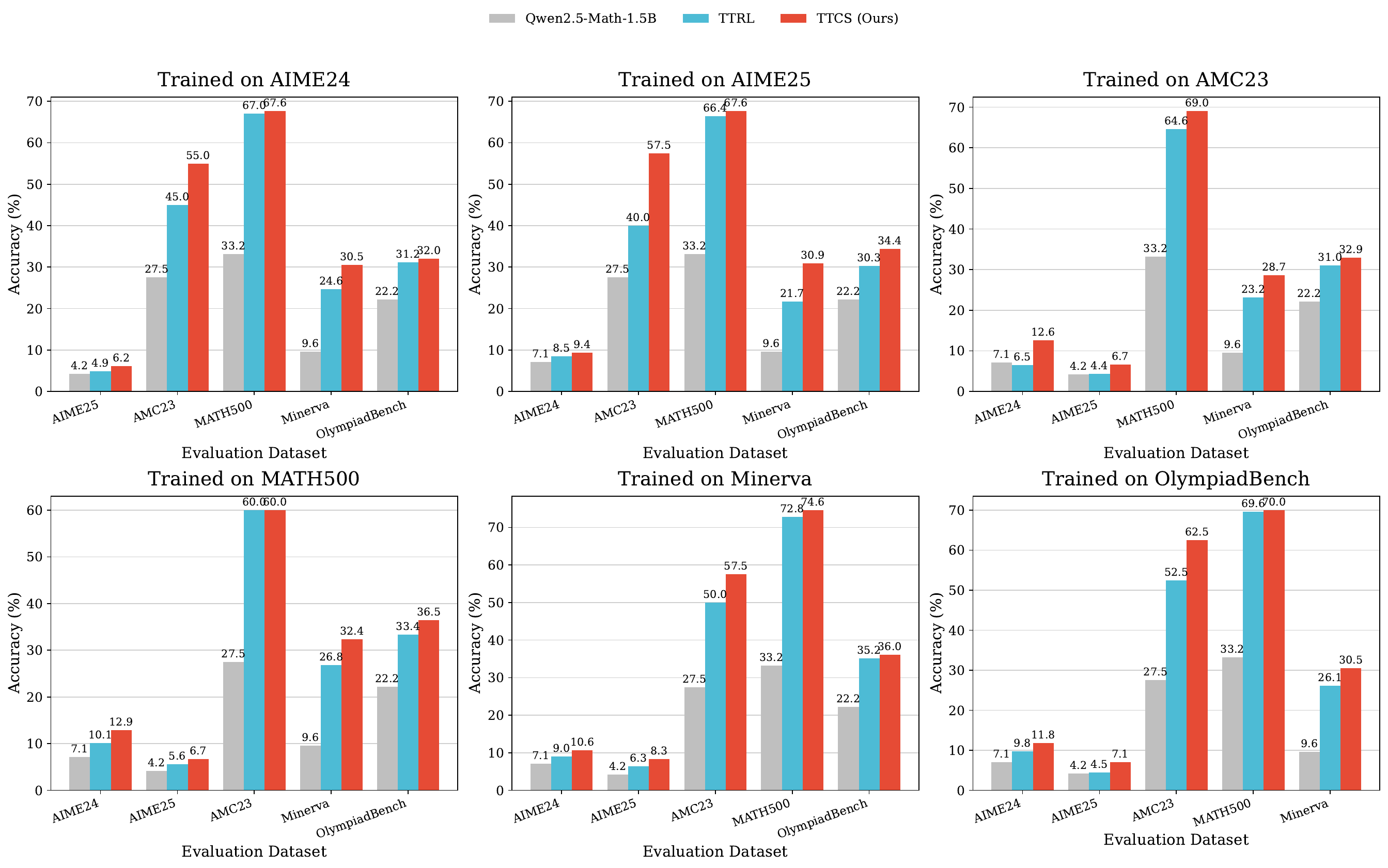} 
    \caption{The out-of-distribution performance comparison of \textbf{TTCS} and other baselines on mathematical benchmarks.}
    \label{fig:appendix-ood-all}
\end{figure*}

To assess the stability of transfer learning, we report the complete out-of-distribution (OOD) evaluation results in Figure~\ref{fig:appendix-ood-all}. Each subplot illustrates the performance trajectories when the model is optimized on a specific source dataset (e.g., AIME24, MATH500) and subsequently evaluated across other distinct mathematical benchmarks. Consistent with our main findings, \sysname{} demonstrates superior robustness, consistently outperforming the TTRL baseline across diverse transfer scenarios. These results confirm that our curriculum-driven approach fosters the acquisition of transferable reasoning skills rather than mere overfitting to the source distribution.

\section{Method Details}
\subsection{Theoretical Analysis: Variance-Driven Generative Synthesis}
\label{appendix:theory_reward}

In this section, given a synthetic question $x'$, we will prove that the bell-shaped capability-adaptive reward $\mathcal{R}_{\mathrm{cap}}(x^\prime) = \big(4\, s(x^\prime) (1 - s(x^\prime))\big)^{\gamma}$ targeting samples with maximal outcome variance when $s = 0.5$ aligns with maximizing the learning signal for the solver.

\paragraph{Gradient Signal and Uncertainty.} Omitting the clipping and KL terms and treating the group-based advantages $A_i$ as constants, the gradient with respect to $\theta$ in the GRPO algorithm(Section ~\ref{subsec:preliminary}) reduces to:
\begin{equation}
    \nabla_\theta \mathcal{J} \propto \mathbb{E}\left[ \sum_{i=1}^G A_i \nabla_\theta \log \pi_\theta(o_i|x') \right].
\end{equation}

The magnitude of the gradient update is modulated by the advantage $A_i$, which means the learning signal strength of the update depends on $\{r_i\}_{i=1}^G$. Since the binary rewards $r_i \in \{0, 1\}$ follow a Bernoulli distribution with parameter $p_\theta(x')$, their variance is $\text{Var}(r_i) = p_\theta(x')(1 - p_\theta(x'))$. When $p_\theta(x') \to 0$ or $1$, the outcomes are nearly deterministic, causing the variance of the advantages to collapse toward zero, which in expectation leads to vanishing gradient updates. Conversely, when $p_\theta(x') \approx 0.5$, the reward variance is maximized, yielding advantage estimates with maximal variance, thereby maximizing the expected magnitude of the stochastic gradient estimator. This specific regime corresponds to the solver's \textit{capability frontier}, where the model is on the verge of mastering a concept but remains unstable.

\paragraph{Self-Consistency as a Proxy.}
Model each rollout as a Bernoulli variable with probability of correctness $p_\theta(x')$. Then $s(x')$ is their sample mean. By Hoeffding’s inequality~\cite{409cf137-dbb5-3eb1-8cfe-0743c3dc925f} for bounded (Bernoulli) variables,
\begin{equation}
    \mathbb{P}\!\left(\big|s(x')-\mathbb{E}[s(x')]\big| \ge \epsilon\right) \le 2 e^{-2G\epsilon^2},
\end{equation}
and under mild conditions (no systematic bias), $\mathbb{E}[s(x')] = p_\theta(x')$. Thus $s(x')$ concentrates exponentially fast around $p_\theta(x')$ as $G$ grows, proving its use as a practical proxy. Consequently, $\mathcal{R}_{\mathrm{cap}}(x^\prime) = \big(4\, s(x^\prime) (1 - s(x^\prime))\big)^\gamma$ targets the high-variance regime, aligning the synthesizer with the solver’s uncertainty frontier.

\paragraph{Practical Implication.}
The KL divergence constraints and clipping to stabilize training in GRPO do not generate learning signals on their own. So the effective magnitude of the policy update is limited by the variance of the advantage estimates, which scales with \(p_\theta(x')(1-p_\theta(x'))\). When \(p_\theta \to 0\) or \(1\), the advantages vanish, rendering the clipping mechanism effectively inactive as the policy ratio variations become too small to hit the trust region boundaries \([1-\epsilon, 1+\epsilon]\). By promoting \(s(x') \approx 0.5\), this capability-adaptive reward steers the synthesizer toward the solver’s capability frontier. This ensures the advantage variance is sufficient to drive meaningful updates that actively engage the clipping mechanism, thereby achieving the optimal balance between efficient exploration and training stability. Finally, in the reward formulation $\mathcal{R}_{\mathrm{cap}}(x^\prime) = \big(4\, s(x^\prime) (1 - s(x^\prime))\big)^\gamma$, the constant \(4\) normalizes the peak to \(1\), while the exponent \(\gamma\) acts as a temperature-like hyperparameter. A larger \(\gamma\) sharpens the focus, aggressively filtering for samples with the highest uncertainty to accelerate learning in the most informative regions.

\subsection{Reference Similarity Penalty}\label{appendix:ref_panalty}
$\mathcal{R}_{\mathrm{ref}}(x_i^\prime, x_{\mathrm{test}})$ is a rule-based metric derived from three similarity scores: text similarity $S_{\mathrm{text}}$, Jaccard similarity $S_{\mathrm{jacc}}$ and skeleton similarity $S_{\mathrm{skel}}$. The penalty is defined hierarchically to prioritize direct textual overlap before checking for structural redundancy:
\begin{equation}
\mathcal{R}_{\mathrm{ref}}(x_i^\prime, x_{\mathrm{test}}) = 
\begin{cases} 
S_{\mathrm{text}}, & \text{if} \quad S_{\mathrm{text}} > \tau_{\mathrm{text}}, \\
S_{\mathrm{skel}}, & \text{elif} \quad (S_{\mathrm{skel}} > \tau_{\mathrm{skel}}) \land \mathcal{C}_{\mathrm{aux}}, \\
0, & \text{otherwise},
\end{cases}
\end{equation}
where $\mathcal{C}_{\mathrm{aux}}$ denotes $(S_{\mathrm{text}} > 0.45 \lor S_{\mathrm{jacc}} > 0.40)$.
$\tau_{text}$ is the adaptive text threshold and $\tau_{skel}$ is the skeleton threshold. Note that structural rejection (the second case) requires auxiliary evidence from either text or Jaccard similarity to prevent false positives.

\subsection{Group Similarity Penalty}\label{appendix:group_panalty}
Following R-Zero~\cite{huang2025rzeroselfevolvingreasoningllm}, we implement a group similarity penalty $\mathcal{R}_{\mathrm{group}}(x^\prime_i, \{x^\prime_{i^{\prime}}\}_{i^{\prime}=1, i^{\prime}\neq i}^{M})$. For each pair of synthetic questions $(x'_p, x'_q)$ in a batch, $x'_p, x'_q \in \{x^\prime_{i^{\prime}}\}_{i^{\prime}=1}^{M}$, we calculate their BLEU-based distance:
\begin{equation}
    d_{p,q} = 1 - \text{BLEU}(x'_p,x'_q).
\end{equation}
Based on $d_{p,q}$, questions are divided into different clusters via agglomerative hierarchical clustering. Considering $x_i'$ in cluster $C_k$, the penalty is defined as:
\begin{equation}
    \mathcal{R}_{\mathrm{group}}(x^\prime_i, \{x^\prime_{i^{\prime}}\}_{i^{\prime}=1, i^{\prime}\neq i}^{M}) = \frac{|C_k|}{B},
\end{equation}
where $|C_k|$ is the number of questions in cluster $C_k$ and $B$ denotes batch size.

\clearpage
\subsection{Algorithm Description}
\label{appendix:TTCS}

\begin{algorithm}[h!]
\caption{\sysname{}: Test-Time Co-Evolution via Iterative GRPO}
\label{alg:TTCS}
\textbf{Require:} base model $\mathcal{M}_0$; unlabeled test set $\mathcal{D}_{\mathrm{test}}$; iterations $T$; \\
\hspace*{1.7em} Synthesizer rollout group size $M$; Synthesizer evaluation sampling $K$; \\
\hspace*{1.7em} Solver sampling size $G$; filtering threshold $\delta$; \\
\hspace*{1.7em} reward hyper-parameters $\gamma, \lambda_1, \lambda_2$; GRPO hyper-parameters $(\epsilon,\beta)$; learning rates $(\eta_\phi,\eta_\theta)$; \\
\textbf{Initialize:} $\pi_\phi^{0}\leftarrow \mathcal{M}_0$, $\pi_\theta^{0}\leftarrow \mathcal{M}_0$
\begin{algorithmic}[1]
\FOR{$t = 1$ \TO $T$}
    \STATE \textbf{// (a) Synthesizer Training: question rollout $\rightarrow$ Solver evaluation $\rightarrow$ GRPO update}
    \STATE Sample test questions $\{x_{\mathrm{test}}^{(b)}\}_{b=1}^{B} \sim \mathcal{D}_{\mathrm{test}}$
    \FOR{$b = 1$ \TO $B$}
        \STATE Roll out auxiliary questions $\{x'_{b,j}\}_{j=1}^{M} \sim \pi_{\phi}^{t-1}(\cdot \mid x_{\mathrm{test}}^{(b)})$
        \FOR{$j = 1$ \TO $M$}
            \STATE Sample Solver responses $\{y_{b,j,i}\}_{i=1}^{K} \sim \pi_{\theta}^{t-1}(\cdot \mid x'_{b,j})$; compute majority vote $\hat y_{\mathrm{maj}}$
            \STATE Compute consistency score $s(x'_{b,j}) \leftarrow \frac{1}{K}\sum_{i=1}^{K}\mathbb{I}[y_{b,j,i}=\hat y]$
            \STATE Compute capability reward $\mathcal{R}_{\mathrm{cap}}(x'_{b,j}) \leftarrow \big(4\,s(x'_{b,j})(1-s(x'_{b,j}))\big)^{\gamma}$
            \STATE Compute similarity penalty $\mathcal{R}_{\mathrm{sim}}(x'_{b,j}) \leftarrow \lambda_1 \mathcal{R}_{\mathrm{ref}}(x'_{b,j},x_{\mathrm{test}}^{(b)}) + \lambda_2 \mathcal{R}_{\mathrm{group}}(x'_{b,j},\{x'_{b,j'}\}_{j'\neq j})$
            \STATE Assign final reward $\mathcal{R}(x'_{b,j}) \leftarrow \mathbb{I}_{\mathrm{valid}}(x'_{b,j})\cdot \max\!\big(0,\;\mathcal{R}_{\mathrm{cap}}(x'_{b,j})-\mathcal{R}_{\mathrm{sim}}(x'_{b,j})\big)$
        \ENDFOR
    \ENDFOR
    \STATE Update Synthesizer by GRPO:
    \STATE \hspace*{1.3em}$\displaystyle \phi^{t} \leftarrow \phi^{t-1} + \eta_{\phi}\,\nabla_{\phi}\,\mathcal{J}_{\mathrm{GRPO}}(\phi)$
    
    \STATE \textbf{// (b) Solver Training: mixed data construction $\rightarrow$ self-supervised reward $\rightarrow$ online filtering $\rightarrow$ GRPO update}
    \STATE Sample test questions $\mathcal{B}_{\mathrm{test}}^{t} \sim \mathcal{D}_{\mathrm{test}}$ (with replacement)
    \STATE Generate curriculum variants $\mathcal{B}_{\mathrm{syn}}^{t}$ by sampling $x' \sim \pi_{\phi}^{t-1}(\cdot \mid x_{\mathrm{test}})$ for each $x_{\mathrm{test}}\in \mathcal{B}_{\mathrm{test}}^{t}$
    \STATE Construct mixed batch $\mathcal{B}^{t}_{\mathrm{train}}\leftarrow \mathcal{B}_{\mathrm{test}}^{t}\cup \mathcal{B}_{\mathrm{syn}}^{t}$
    \STATE Initialize filtered batch $\widetilde{\mathcal{B}}^{t}\leftarrow \emptyset$
    \FOR{each $x \in \mathcal{B}^{t}_{\mathrm{train}}$}
        \STATE Sample a response group $\{y_i\}_{i=1}^{G} \sim \pi_{\theta}^{t-1}(\cdot \mid x)$; compute consensus $\hat{y}^*$
        \STATE Assign rewards $r_i \leftarrow \mathbb{I}[y_i=\hat{y}^*]$ and consistency score $s(x)\leftarrow \frac{1}{G}\sum_{i=1}^{G} r_i$
        \IF{$|s(x)-0.5|\le \delta$} \STATE $\widetilde{\mathcal{B}}^{t}\leftarrow \widetilde{\mathcal{B}}^{t}\cup\{(x,\{y_i\},\{r_i\})\}$ \ENDIF
    \ENDFOR
    \STATE Update Solver by GRPO:
    \STATE \hspace*{1.3em}$\displaystyle \theta^{t} \leftarrow \theta^{t-1} + \eta_{\theta}\,\nabla_{\theta}\,\mathcal{J}_{\mathrm{GRPO}}(\theta)$

\ENDFOR
\STATE \textbf{return} $\pi_{\phi}^{T}, \pi_{\theta}^{T}$
\end{algorithmic}
\end{algorithm}

\section{Synthetic Prompt Template}\label{appendix:prompt}
\newtcolorbox{promptbox}[2][]{%
    enhanced,              
    breakable,             
    colback=gray!5,        
    colframe=gray!75!black,
    coltitle=white,        
    fonttitle=\bfseries\sffamily, 
    title={#2},            
    attach boxed title to top left={xshift=5mm, yshift*=-\tcboxedtitleheight/2}, 
    boxed title style={    
        colback=gray!75!black,
        sharp corners, 
        rounded corners=northeast, 
        rounded corners=southeast,
        boxrule=0pt,
    },
    boxrule=0.5pt,         
    sharp corners=south,   
    arc=3mm,               
    top=12pt,              
    fontupper=\ttfamily\small, 
    #1                     
}

\newcommand{\var}[1]{\textcolor{blue}{\textbf{\{\{#1\}\}}}}

Here we present the full prompt template used for the self-evolution process.
\begin{promptbox}{Prompt Template: Isomorphic Problem Generator}
You are an expert mathematics problem setter. Your task is to generate a \textbf{**Structurally Isomorphic**} but \textbf{**Surface-Distinct**} problem based on a provided "Reference Question."

\textbf{Instructions:}

1. \textbf{Analyze \& Isolate:} Identify the "Decisive Lemma" (the single core logical step) required to solve the reference.

2. \textbf{Object Mapping \& Structural Shift:}
\begin{itemize}
    \item Map the reference's variables/setting to a different mathematical representation.
    \item \textbf{Constraint:} At least one of the \textbf{Main Objects}, \textbf{Setting}, or \textbf{Constraint Type} must change. This must change \textit{what is being counted/optimized/constructed}, not merely rename variables or swap story nouns.
\end{itemize}

3. \textbf{Immediate-Equivalence Embargo (CRITICAL):}
\begin{itemize}
    \item The new problem statement must \textbf{NOT} explicitly include any immediate algebraic equivalent of the reference's key identity.
    \item \textbf{Banned forms:} Expanded polynomials, completing the square, Vieta restatements, or absolute-value splits \textbf{IF AND ONLY IF} they make the reference skeleton immediately obvious.
    \item \textbf{Allowed:} Standard mathematical notation (like modulo arithmetic or standard substitutions) is permitted \textbf{ONLY IF} it is necessary for a natural, concise problem statement and does not trivially reveal the solution path.
\end{itemize}

4. \textbf{Verifiable Complexity Alignment:}
\begin{itemize}
    \item Ensure the reduced search space is verifiable and comparable.
    \item In the Design block, you must state exactly one \textbf{Concrete Metric}.
    \item \textbf{Selection Rule:} Choose the metric that best reflects the \textbf{dominant search/optimization dimension}. Do not default to \texttt{key-steps} unless no other option fits.
    \item \textbf{Formats:}
    \begin{itemize}
        \item \texttt{\#cases=k} (Enumeration/Casework)
        \item \texttt{\#factor-pairs=k} (Number Theory)
        \item \texttt{DOF=n} (Geometry/Optimization degrees of freedom)
        \item \texttt{independent-axes=k} (Combinatorics/Probability dimensions)
        \item \texttt{key-steps=k} (Fallback for logic/recursion depth, integer \$k \textbackslash in [2, 8]\$)
    \end{itemize}
    \item \textbf{Constraint:} Do not use vague terms like "similar magnitude".
\end{itemize}

5. \textbf{Output Type Lock (CRITICAL):}
\begin{itemize}
    \item The Final Answer must be a \textbf{Single Scalar Number} (Integer, Fraction, or Simplest Radical Form).
    \item \textbf{Aggregation Rule:} If the math naturally yields a set of solutions (e.g., "Find all \$x\$"), you MUST change the question to ask for the \textbf{Sum}, \textbf{Product}, \textbf{Count}, or \textbf{Maximum} of those solutions.
    \item \textbf{Forbidden Outputs:}
    \begin{itemize}
        \item \textbf{No Sets:} (e.g., \$\textbackslash\{\{1, 2\}\textbackslash\}\$ is banned \$\textbackslash to\$ Ask for \$1+2=3\$).
        \item \textbf{No Functions:} (e.g., \$f(x)=e\textasciicircum x\$ is banned \$\textbackslash to\$ Ask for \$f(1)\$).
        \item \textbf{No Text/Boolean:} (e.g., "Yes", "True", "Convergent" are banned).
        \item \textbf{No Proofs:} Never ask "Prove that...".
        \item \textbf{No Subparts:} The question must be atomic (no "(a)... (b)...").
    \end{itemize}
\end{itemize}

6. \textbf{Fail-Safe Rule:}
\begin{itemize}
    \item If you cannot ensure isomorphism without leaking the skeleton or losing uniqueness, \textbf{discard the current strategy} and switch to a different Shift Strategy (e.g., Domain Transfer \$\textbackslash leftrightarrow\$ Algebraic Masking) before generating.
\end{itemize}

\textbf{Output Format (Strict):}

1. \textbf{[Design]:}
\begin{itemize}
    \item \textit{Lemma:} One short sentence.
    \item \textit{Shift:} Strategy used (e.g., "Divisibility \$\textbackslash to\$ Equation").
    \item \textit{Verification:} Confirm uniqueness and include the \textbf{Concrete Metric}. Do NOT reveal the specific reduced equation.
\end{itemize}
2. \textbf{\textless question\textgreater Block:}
\begin{itemize}
    \item Start immediately with \texttt{\textless question\textgreater}.
    \item Content: The problem statement in LaTeX. \textbf{NO} headings, \textbf{NO} introductions.
    \item End immediately with \texttt{\textless /question\textgreater}.
\end{itemize}
3. \textbf{Final Answer:}
\begin{itemize}
    \item Must be the \textbf{last line}.
    \item Format strictly: \texttt{Final Answer: \textbackslash boxed\{\{value\}\}} (Value must be a single scalar number).
\end{itemize}

\textbf{Examples:}

\textit{[... Few-shot examples demonstrating domain transfer (Number Theory \$\textbackslash to\$ Geometric Combinatorics), algebraic masking, and complex analysis mappings are omitted for brevity ...]}
---

\textbf{Current Task:}
\textbf{Reference Question:}
\var{reference\_question}

\textbf{[Design]:}
\end{promptbox}

\newpage
\section*{NeurIPS Paper Checklist}

\begin{enumerate}

\item {\bf Claims}
    \item[] Question: Do the main claims made in the abstract and introduction accurately reflect the paper's contributions and scope?
    \item[] Answer: \answerYes{} 
    \item[] Justification: The claims in the abstract and introduction accurately reflect the proposed method and are supported by empirical results.

\item {\bf Limitations}
    \item[] Question: Does the paper discuss the limitations of the work performed by the authors?
    \item[] Answer: \answerNo{} 
    \item[] Justification: Due to space constraints, the current submission does not provide a separate discussion of limitations.

\item {\bf Theory assumptions and proofs}
    \item[] Question: For each theoretical result, does the paper provide the full set of assumptions and a complete (and correct) proof?
    \item[] Answer: \answerNo{} 
    \item[] Justification: The paper focuses on empirical methods and does not include formal theoretical results.

    \item {\bf Experimental result reproducibility}
    \item[] Question: Does the paper fully disclose all the information needed to reproduce the main experimental results of the paper to the extent that it affects the main claims and/or conclusions of the paper (regardless of whether the code and data are provided or not)?
    \item[] Answer: \answerYes{} 
    \item[] Justification: We provide sufficient details on data, models, and procedures to reproduce the main results.

\item {\bf Open access to data and code}
    \item[] Question: Does the paper provide open access to the data and code, with sufficient instructions to faithfully reproduce the main experimental results, as described in supplemental material?
    \item[] Answer: \answerYes{} 
    \item[] Justification: Code is released in anonymized form with instructions for reproducing the results.

\item {\bf Experimental setting/details}
    \item[] Question: Does the paper specify all the training and test details (e.g., data splits, hyperparameters, how they were chosen, type of optimizer) necessary to understand the results?
    \item[] Answer: \answerYes{} 
    \item[] Justification: All key experimental settings, including hyperparameters and evaluation protocols, are described.

\item {\bf Experiment statistical significance}
    \item[] Question: Does the paper report error bars suitably and correctly defined or other appropriate information about the statistical significance of the experiments?
    \item[] Answer: \answerNo{} 
    \item[] Justification: Due to computational cost, experiments are conducted with limited runs; we instead validate consistency across benchmarks.

\item {\bf Experiments compute resources}
    \item[] Question: For each experiment, does the paper provide sufficient information on the computer resources (type of compute workers, memory, time of execution) needed to reproduce the experiments?
    \item[] Answer: \answerYes{} 
    \item[] Justification: We report the compute resources and training costs used in experiments.
    
\item {\bf Code of ethics}
    \item[] Question: Does the research conducted in the paper conform, in every respect, with the NeurIPS Code of Ethics \url{https://neurips.cc/public/EthicsGuidelines}?
    \item[] Answer: \answerYes{} 
    \item[] Justification: The research complies with the NeurIPS Code of Ethics and uses public data.

\item {\bf Broader impacts}
    \item[] Question: Does the paper discuss both potential positive societal impacts and negative societal impacts of the work performed?
    \item[] Answer: \answerNo{} 
    \item[] Justification: The paper focuses on methodological contributions and does not explicitly discuss societal impacts; however, no immediate harmful applications are identified.
    
\item {\bf Safeguards}
    \item[] Question: Does the paper describe safeguards that have been put in place for responsible release of data or models that have a high risk for misuse (e.g., pre-trained language models, image generators, or scraped datasets)?
    \item[] Answer: \answerNA{} 
    \item[] Justification: No high-risk assets are released.

\item {\bf Licenses for existing assets}
    \item[] Question: Are the creators or original owners of assets (e.g., code, data, models), used in the paper, properly credited and are the license and terms of use explicitly mentioned and properly respected?
    \item[] Answer: \answerYes{} 
    \item[] Justification: All datasets and models are properly cited and used in accordance with their licenses.

\item {\bf New assets}
    \item[] Question: Are new assets introduced in the paper well documented and is the documentation provided alongside the assets?
    \item[] Answer: \answerNA{} 
    \item[] Justification: No new datasets or models are released.

\item {\bf Crowdsourcing and research with human subjects}
    \item[] Question: For crowdsourcing experiments and research with human subjects, does the paper include the full text of instructions given to participants and screenshots, if applicable, as well as details about compensation (if any)? 
    \item[] Answer: \answerNA{} 
    \item[] Justification: No human subjects or crowdsourcing are involved.

\item {\bf Institutional review board (IRB) approvals or equivalent for research with human subjects}
    \item[] Question: Does the paper describe potential risks incurred by study participants, whether such risks were disclosed to the subjects, and whether Institutional Review Board (IRB) approvals (or an equivalent approval/review based on the requirements of your country or institution) were obtained?
    \item[] Answer: \answerNA{} 
    \item[] Justification: Not applicable as no human subjects are involved.

\item {\bf Declaration of LLM usage}
    \item[] Question: Does the paper describe the usage of LLMs if it is an important, original, or non-standard component of the core methods in this research? Note that if the LLM is used only for writing, editing, or formatting purposes and does \emph{not} impact the core methodology, scientific rigor, or originality of the research, declaration is not required.
    \item[] Answer: \answerNA{} 
    \item[] Justification: LLMs are used only for writing and editing purposes and are not part of the core methodology.

\end{enumerate}

\end{document}